\documentclass[twoside]{article}
\usepackage{ecj,palatino,epsfig,latexsym} 
\usepackage[sort]{natbib}

\usepackage{pdflscape}
\usepackage{amsthm,amsfonts,amsmath,amssymb,bm,bbm}
\usepackage{dsfont}
\usepackage{hyperref}
\usepackage{color}
\usepackage{subfigure}
\usepackage{rotating}
\usepackage{xspace}
\usepackage{wrapfig}
\usepackage{floatpag}
\usepackage{multirow}
\usepackage{algorithm}
\usepackage{rotating}
\usepackage{tikz}
\usepackage{graphicx}

\graphicspath{{/plots/}{plots/}}
\newcommand{\ignore}[1]{}

\newcommand{\tmax}{t_{\max}}

\newcommand{\rwalk}{EA-UniformWalk\xspace}
\newcommand{\bwalk}{EA-BiasedWalk\xspace}
\newcommand{\arwalk}{EA-AsymUniformWalk\xspace}
\newcommand{\abwalk}{EA-AsymBiasedWalk\xspace}

\newcommand{\ONEMAX}{\textsc{OneMax}\xspace}

\newcommand{\plotyellow}{\includegraphics[height=1.7mm]{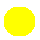}}
\newcommand{\plotrot}{\includegraphics[height=1.7mm]{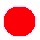}}
\newcommand{\plotlime}{\includegraphics[height=1.7mm]{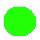}}
\newcommand{\plotnavy}{\includegraphics[height=1.7mm]{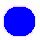}}
\newcommand{\plotmagenta}{\includegraphics[height=1.7mm]{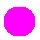}}
\newcommand{\plotcyan}{\includegraphics[height=1.7mm]{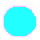}}
\newcommand{\plotblack}{\includegraphics[height=1.7mm]{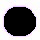}}
\newcommand{\brw}{biased random walk\xspace}
\newcommand{\rw}{random walk\xspace}
\newcommand{\Hue}{\mbox{\em{Hue}}}
\newcommand{\Col}{\mbox{\em{Color}}}

\newcommand{\Ben}{\mbox{\em{Ben}}}
\newcommand{\GCF}{\mbox{\em{GCF}}}
\newcommand{\lc}{\mbox{\em{lc}}}
\newcommand{\lum}{\mbox{\em{lum}}}
\usepackage{todonotes}

\begin{document}

\title{\bf Evolutionary Image Transition and Painting Using Random Walks}

\author{\name{\bf Aneta Neumann} \hfill 
\addr{aneta.neumann@adelaide.edu.au}\\
        \addr{Optimisation and Logistics, School of Computer Science, The University of Adelaide, Australia}     
\AND
\name{\bf Bradley Alexander} \hfill \addr{bradley.alexander@adelaide.edu.au}\\
        \addr{Optimisation and Logistics, School of Computer Science, The University of Adelaide, Australia}
\AND
\name{\bf Frank Neumann} \hfill \addr{frank.neumann@adelaide.edu.au}\\
        \addr{Optimisation and Logistics, School of Computer Science, The University of Adelaide, Australia}
}

\maketitle
\begin{abstract}
We present a study demonstrating how random walk algorithms can be used for evolutionary image transition.
We design different mutation operators based on uniform and biased random walks and study how their combination with a baseline mutation operator can lead to interesting image transition processes in terms of visual effects and artistic features. Using feature-based analysis we investigate the evolutionary image transition behaviour with respect to different features and evaluate the images constructed during the image transition process.
Afterwards, we investigate how modifications of our biased random walk approaches can be used for evolutionary image painting.
We introduce an evolutionary image painting approach whose underlying biased random walk can be controlled by a parameter influencing the bias of the random walk and thereby creating different artistic painting effects.
\end{abstract}

\sloppy

\section{Introduction}\label{sec:intro}

Evolutionary algorithms (EAs) have been widely and successfully applied in the areas of art~\citep{DBLP:conf/ncs/2008evolution,hingston2008design,mccormack2012computers,DBLP:conf/siggraph/LambertLL13,DBLP:journals/alife/AntunesLL15,DBLP:conf/gecco/NeumannN18,DBLP:conf/gecco/Neumann019}. In this application area, the primary aim is to evolve artistic and creative outputs through an evolutionary process~\citep{citeulike:12541313,DBLP:conf/evoW/VinhasACEM16,DBLP:conf/evoW/al-RifaieB13a,DBLP:conf/evoW/Greenfield15,DBLP:conf/gecco/NeumannSCN17,DBLP:conf/cec/0001N18}. 
The use of evolutionary algorithms for the generation of art has attracted strong research interest. 
Different representations have been used to create works of greater complexity in 2D and 3D~\citep{todd1992evolutionary,DBLP:journals/alife/GreenfieldM15,DBLP:conf/gecco/MachadoC14}, and in image animation~\citep{DBLP:conf/siggraph/Sims91,DBLP:conf/evoW/Hart07,DBLP:journals/ijart/TristCB11,DBLP:journals/corr/abs-1710-07421}.  
The great majority of this work relates to the use of evolution to produce a final artistic product in the form of a picture, sculpture, animation.

\subsection{Related work}

In the seminal work the \emph{Blind Watchmaker},~\cite{dawkins1986blind} investigated the process of evolution. He evolved figures called biomorphs in order to explore and to visualize the power of evolution. The biomorphs' appearance changes/diverges broadly from the original parent over time demonstrating a simple model of bio-inspired evolution.
Inspired by Dawkins' work,~\cite{DBLP:conf/siggraph/Sims91} created complex and imaginative images. He used an expression-based approach to evolve images using a mathematical expression as the genotype and applied crossover and mutation. In the interactive media installation \emph{Galapagos}, visitors were able to evolve virtual creatures based on Darwinian evolution principles whilst taking into account their own preferences~\citep{sims}. The abstract organisms were displayed on computer screens and users interactively chose virtual organisms for simulated growth by stepping onto selected pads. This exhibition was a visualization example of a collaboration process between the visitors and the machine.

\cite{latham1985} created a black and white lithographic artwork called \emph{Black Form Synth} consisting of hand-drawn "evolutionary trees of complex forms" in the form of a flow diagram arranged with a set of transformation rules. Following the expression-based approach,~\cite{synth_latham} extended the rule-based evolutionary approach in order to invent new complex 3D forms from geometric primitives. \cite{todd1992evolutionary} introduced the framework called \emph{Mutator} to generate art and evolve new 3D biomorphic forms. The \emph{Mutator} creates a series of complex branching organic forms and surreal virtual sculptures, and animated videos through the process of "surreal" evolution. At each iteration the artist selects phenotypes based on the idea of evolved organisms that are able to ``breed and grow''.

Over time, the growing interest in an expression-based approach led to expansion of the area of evolutionary art, providing coherent and solid basis for further research~\citep{DBLP:conf/gecco/FentonMFFHO17,10.1007/978-3-319-77583-8_1,10.1007/978-3-319-77583-8_19,10.1007/978-3-030-16667-0_6}.~\cite{DBLP:conf/evoW/Hart07} used an expression-based approach with the focus of evolving a set of images with a different appearance to the previous works. This system's interface allowed a greater range of control over the colors and forms of the evolved images. Complex and detailed images were created in~\cite{rooke} using an expression-based approach guided by aesthetic selection, in particular towards the evolution of color space.
~\cite{DBLP:conf/smc/Unemi04,DBLP:conf/cec/Unemi12} continued to explore the expression-based approach to evolutionary art by evolving images and animations towards the progression of color volumes and novel forms with varied variables. The images are evolved based on aesthetic measures and have been displayed on the internet for decades using different versions of the SBART framework~\citep{unemi1999sbart}.~\cite{DBLP:journals/apin/MachadoC02} introduced computer-aided software, called \emph{NEvAr}. This is an evolutionary art tool which uses: genetic programming, user-guided evolution, and automatic fitness assignment. The model uses a fitness function that permits aesthetically pleasing and visually complex images.
~\cite{draves2005electric} introduced \emph{Electric Sheep}, a large and ongoing evolutionary art project using collective human evaluation. It was implemented as a distributed screen-saver allowing users to approve or disapprove phenotypes in order to evolve artificial life and is focused on the continuing behavior of the distributed system. ~\cite{greenfield2006robot} describes simulated robots embedded into an evolutionary framework for the purpose of painting a new artwork. The system implements optimization towards identifying robot paintings with higher aesthetic properties and takes into account the behavior of the simulated robots.

Aesthetic feature measures have been widely applied to generate new artistic images~\citep{lewis2008evolutionary,DBLP:conf/iconip/NeumannPA18,10.1007/s10710-018-9336-1}. The evolutionary art system \textit{NEvAr}, \cite{DBLP:journals/apin/MachadoC02} used aesthetic measures and evolutionary computation methods, and thus developed automatic seeding procedures to generate images. \cite{DBLP:journals/ijcia/Greenfield02} evolved images using computational aesthetic functions that are based on a color segmentation algorithm. Moreover, \cite{DBLP:journals/swevo/HeijerE14} studied aesthetic measures in unsupervised evolutionary art using aesthetic measures as fitness functions.

In \emph{Interactive Evolutionary Computation} (IEC), the traditional objective-function is replaced by a human subject who guides the selection~\citep{takagi2001interactive}. In IEC, each solution is evaluated by a human judgment. Those subjective choice provides the basis for the selection of solutions in the evolutionary process. IEC founds its biggest application in the creative domains in which including humans `in the loop’ can lead to novel solutions~\citep{takagi2001interactive,DBLP:conf/ncs/2008evolution,hingston2008design,Fenton:2017:PGE:3067695.3082469,DBLP:conf/cec/HollingsworthS19}. In this case, the fitness function is based on the individual user's experiences and preferences towards interesting or aesthetic results. This means that the fitness of the evolutionary algorithm (EA) is strongly subjective.

In contrast to this, a different approaches evaluate artistic images by automate fitness evaluation in order to generate novel images. \cite{DBLP:journals/connection/BalujaPJ94} attempted to automate the process of image evolution by using neural networks in order to produce images. The primary idea was to learn the user's preferences, and to apply this insight to generate aesthetically pleasing images. 
Moreover,~\cite{DBLP:journals/swevo/HeijerE14} presented an approach for evolving images without human interaction. Instead of human interaction, they used one or more aesthetic measures as fitness functions to guide the search. They investigated the correlation between aesthetic scores in order to calculate which aesthetic measure generate images that are evaluated positively by the other aesthetic measures.

Another application of evolutionary algorithms to art is the creation of image transitions. 
~\cite{DBLP:conf/eps/GrafB95} used interactive evolution to help determine parameters for image morphing. They combined interactive evolutionary computation with the concept of warping and morphing from computer graphics to evolve images.
More precisely, they used the recombination of two bitmap images through image interpolation. Furthermore,~\cite{karungaru2007automatic} 
used an evolutionary algorithm to automatically identify features for morphing faces.

More recently, deep neural networks have been used to create artistic images through the transfer of artistic style from one image to another, facilitating novel forms of image manipulation~\citep{gatys2016image}. A new neural style transfer approach preserves the colors of the original image using simple linear methods for transferring style.~\cite{gatys2016preserving,gatys2017controlling} extended
the existing method by introducing control over colour information, spatial scale, and spatial location.
\cite{johnson2016perceptual} combined the benefits of both approaches, trained feed-forward convolutional neural networks and perceptual loss functions based on high-level features extracted from pre-trained networks and proposed the use of perceptual loss functions for training feed-forward networks for image transformation. The transfer of style from one image to stable video sequences was proposed by new initializations and loss functions applicable to videos in~\cite{ruder2016artistic}.

Furthermore, inspired by the work of \cite{DBLP:conf/ppsn/GaoNN16}, evolutionary diversity optimization have been applied to evolve images in one/two aesthetic feature dimensions~\citep{DBLP:conf/gecco/AlexanderKN17,DBLP:conf/gecco/NeumannGDN018,DBLP:conf/iconip/AlexanderHNU19}. Building on these studies,  \cite{DBLP:conf/gecco/NeumannG0019} introduced evolutionary diversity optimization for images using popular indicators from the area of evolutionary multi-objective optimisation.

Non-photorealistic rendering (NPR) is an area of computer graphics that focuses on enabling a wide variety of expressive styles for digital art~\citep{Strothotte:2002:NCG:544522}. In contrast to traditional computer graphics, which has focused on photorealism, NPR is inspired by artistic styles such as painting, drawing, technical illustration, and animated cartoons.
~\cite{Litwinowicz:1997:PIV} described transformations from ordinary video segments into animations that appear similar to hand-painted techniques. In particular, they used modified off-the-shelf image processing and rendering techniques in order to process images and videos for an impressionist effect.
~\cite{Kubelka:48} introduced the optical properties model in order to simulate the optical effect of different layers of artwork. \cite{DBLP:conf/siggraph/Hertzmann98} generated hand-painted images by using series of different brush stroke sizes and orientations to generate images.

In recent years, several evolutionary approaches for the production of non-photorealistic renderings of images have been introduced~\citep{izadi2010evolved,barile2008non,kang2005multi,DBLP:conf/gecco/MachadoP12,journals/cin/Wu18}.

Furthermore, extensive research has been carried out on swarm painting.~\cite{DBLP:conf/evoW/Urbano06} investigated choices inside a group of agents in  order to design swarm art with interesting random  patterns.~\cite{DBLP:conf/ecal/MonmarcheSV99} used the stochastic and exploratory principles of an ant colony to automatically discover clusters in data without prior knowledge of the structure of the data.

A swarm-based system was used as a method to create visualizations of data, and in order to combine information aesthetics with data visualization~\citep{DBLP:conf/ijcai/MacasCMM15}.~\cite{Boyd:2004:SIA:1027527.1027674} created a collaborative project called \textit{Swarm Art} that incorporated the swarm-based simulation and projected the artwork onto a large screen. ~\cite{DBLP:conf/evoW/Greenfield05} used a colony optimization model to evolve ant paintings and investigated the effects of different fitness measures on the generation of different artistic styles. Inspired by natural phenomena, namely the use of pheromone substances for mass recruitment in ants' pattern behaviour,~\cite{DBLP:conf/evoW/Urbano05} generated collective artistic work by embedding a pheromone medium pattern into the painters' behaviour.

\subsection{Our Contribution}

~\cite{DBLP:conf/iconip/NeumannAN16} described an image transition process where the key idea is to use the evolutionary process {\em{itself}} in an artistic way.
The focus of our paper is to study how random walk algorithms can be used in the evolutionary image transition process defined in~\cite{DBLP:conf/iconip/NeumannAN16} as mutation operators. 
We consider the well-studied (1+1)~EA, popular random walk algorithms and provide a
new approach to evolutionary art by using theoretical approaches for evolutionary image transition.
The transition process consists of evolving a given starting image $S$ into a given target image $T$ by random decisions. Considering an error function which assigns to a given current image $X$ the number of pixels where it agrees with $T$ and maximizes this function boils down to the classical \ONEMAX problem for which numerous theoretical results on the runtime behaviour of evolutionary algorithms are available~\citep{DBLP:journals/ec/JansenS10,DBLP:journals/cpc/Witt13,DBLP:journals/tec/Sudholt13}. An important topic related to the theory of evolutionary algorithms are random walks~\citep{Lovasz1996,Dembo2004}. We consider random walks on images where each time the walk visits a pixel its value is set to the value of the given target image. By biasing the random walk towards pixels that are similar to the current pixel we can study the effect of such biases which might be more interesting from an artistic perspective.
After observing these two basic random processes for image transition, we study how they can be combined to give the evolutionary process interesting new properties. We study the effect of running random walks for short periods of time as part of a mutation operator in a (1+1) EA. Furthermore, we consider the effect of combining them with the asymmetric mutation operator for evolutionary image transition introduced in~\cite{DBLP:conf/iconip/NeumannAN16}. Our results show that the area of evolutionary image transition based on random walks provides a rich source of artistic possibilities for creating video art. All our approaches are pixel-based and created videos based on the evolutionary processes show frames corresponding to images that were created every few hundred generations. After introducing these different approaches to evolutionary image transition based on random walks, we study their behaviour with respect to different aesthetic features. 
Feature-based analysis of heuristic search methods has gained increasing interest in recent years~\citep{DBLP:conf/ppsn/MersmannPT10,DBLP:journals/amai/MersmannBT0BN13,DBLP:conf/foga/Nallaperuma0NBMT13}. In other application areas feature-based analysis is an important method to increase the theoretical understanding algorithm performance and particularly useful for algorithm selection and configuration~\citep{DBLP:journals/firai/Nallaperuma0N15,DBLP:journals/corr/PoursoltanN16}.
For evolutionary image transition, we study how artistic features behave during the transition process. This allows the measurement of the evolutionary image transition process in a quantitative way and provides a basis to compare our different approaches with respect to artistic measures~\footnote{Images and videos are available at \url{https://vimeo.com/anetaneumann}}. 
\begin{figure}[t] 
\vspace{0.1cm}
\centering
\vskip -6pt
\vspace{0.1cm}
\includegraphics[scale=0.198]{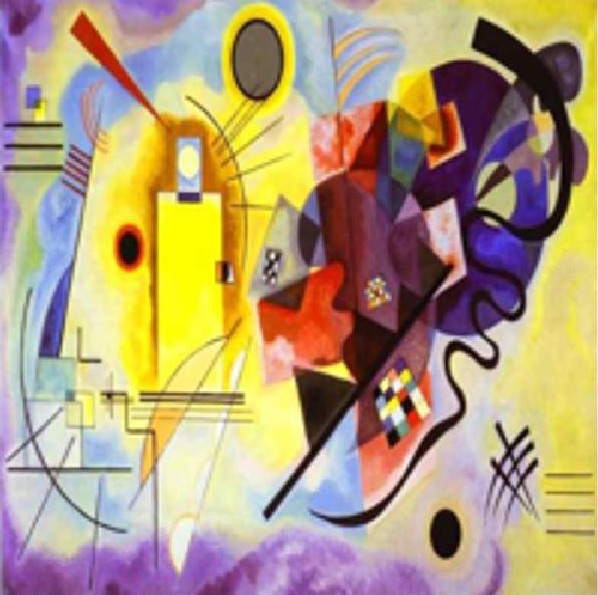}
\includegraphics[scale=0.235]{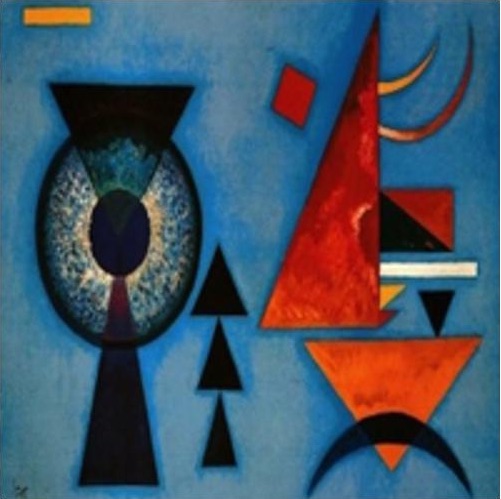}\\
\vspace{0.1cm}
\caption{Starting image X (Yellow-Red-Blue, 1925 by Wassily Kandinsky) and target image T (Soft Hard, 1927 by Wassily Kandinsky).}
\label{fig:2}
\end{figure}

This article extends its conference version~\citep{NeumannEvo2017} in several ways. Firstly, by investigations on the impact of random walk lengths for image transition in Section 4.2 and the impact of the bias of the random walks controlled by the parameter $\alpha$ in Section 4.3. Large values of $\alpha$ increase the probability of moving to similar color pixel and different values of $\alpha$ lead to different image transition processes.
Secondly, we investigate how random walk algorithms can be used to carry out painting of images in Section 7. The key idea is to use a biased random walk starting at a given pixel and color all pixels visited by the walk with the color of the starting pixel. We use the biased random walk mutation approach introduced for evolutionary image transition and combine it with the asymmetric mutation operator. Our approach starts a biased random walk for each pixel to be changed by asymmetric mutation. To achieve different effects of evolutionary image painting, we consider the parameter $\alpha$ which allows to control of the bias of the random walk towards similar color pixels. Our experimental results show the effect of the setting of $\alpha$ for images of various types.

The outline of the paper is as follows. In Section~\ref{sec2}, we introduce the evolutionary transition process. In Section~\ref{sec3}, we study how variants of random walks can be used for the image transition process. We examine the use of random walks as part of mutation operators and study their combinations with asymmetric mutation during the evolutionary process in Section~\ref{sec4}. Furthermore, we extend our investigation on the impact of different random walk length in biased random walk mutation and on the impact of the chosen $\alpha$ in biased random walk mutation.
In Section~\ref{sec6}, we analyse the different approaches for evolutionary image transition with respect to aesthetic features. 
Our evolutionary painting algorithm using biased random walks is introduced and evaluated in Section~\ref{sec7}. Finally, we finish with some conclusions and future work.

\section{Evolutionary Image Transition}\label{sec2}

\begin{algorithm}[t]
\vspace{0.1cm}
\begin{itemize} 
\item Let $S$ be the starting image and $T$ be the target image.
\item Set X:=S.

\item Evaluate $f(X,T)$.
\item while (not termination condition)
\begin{itemize}
\item Obtain image $Y$ from $X$ by mutation.
\item Evaluate $f(Y, T)$
\item If $f(Y, T) \geq f(X, T)$, set $X:= Y$. 
\end{itemize}
\end{itemize}
\caption{(1+1)~EA for evolutionary image transition}
\label{alg:ea}

\end{algorithm}

We consider the evolutionary image transition process introduced in~\cite{DBLP:conf/iconip/NeumannAN16}.
It transforms a given image $S=(S_{ij})$ of size $m \times n$ into a given target image $T= (T_{ij})$ of size $m \times n$. This is done by producing images $X$ for which $X_{ij} \in \{S_{ij}, T_{ij} \}$ holds.
Given a starting image $S=(S_{ij})$ a target image $T=(T_{ij})$, and a current image $X= (X_{ij})$,
we say that pixel $X_{ij}$ is in state $s$ if $X_{ij} = S_{ij}$, and $X_{ij}$ is in state $t$ if $X_{ij}=T_{ij}$.
Our goal is to study different ways of using random walk algorithms for evolutionary image transition.

Throughout this paper, we assume that $S_{ij} \not = T_{ij}$ as pixels with $S_{ij} = T_{ij}$ can not change values and therefore do not have to be considered in the evolutionary process.
To illustrate the effect of the different methods presented in this paper, we consider the work Yellow-Red-Blue, 1925 by Wassily Kandinsky as the starting image and the work T Soft Hard, 1927 by Wassily Kandinsky as the target image (see Figure~\ref{fig:2}). In principle, this process can be carried out with any starting and target image. Using artistic images in this setting has the advantage that artistic properties of images are transformed during the evolutionary image transition process. We will later on in this paper study how the different operators used in the algorithms influence artistic appearance in respect to different artistic features.
At the beginning of the 20th century the well-known artist Wassily Kandinsky was part of the famous Bauhaus movement~\citep{droste2002bauhaus}. Kandinsky was a unique university teacher in Weimar and Dessau and an iconic promoter of 
a theory of geometric figures and their relationships. In the work "Point and Line to Plane",~\cite{kandinsky1979point} show an innovative approach to artistic expression and to the creation of abstract paintings.

We use the fitness function for evolutionary image transition used in~\cite{DBLP:conf/iconip/NeumannAN16} and measure the fitness of an image $X$ as the number of pixels where $X$ and $T$ agree. This fitness function is equivalent to the \ONEMAX problem when interpreting the pixels of $S$ as $0$'s and the pixels of $T$ as $1$'s. Hence, the fitness of an image $X$ with respect to the target image $T$ is given by
\[
f(X,T) = |\{X_{ij} \in X \mid X_{ij}=T_{ij}\}|.
\]

We consider simple variants of the classical (1+1)~EA in the context of image transition. The algorithm is using mutation only and accepts an offspring if it is at least as good as its parent according to the fitness function. The approach is given in Algorithm~\ref{alg:ea}.
Using this algorithm has the advantage that the parent and offspring only differ by a small number of pixels which allows a smooth transition process. This ensures a smooth process for transitioning the starting image into the target. Furthermore, we can interpret each step of the random walks flipping a visited pixel to the target outlined in Section~\ref{sec3} as a mutation step which, according to the fitness function, is always accepted.

\begin{algorithm}[t]
\vspace{0.1cm}
\begin{itemize}
\item Obtain $Y$ from $X$ by flipping each pixel $X_{ij}$ of $X$ independently of the others with probability $c_s/(2|X|_S)$ if $X_{ij}=S_{ij}$, and flip $X_{ij}$ with probability $c_t/(2|X|_T)$ if $X_{ij}=T_{ij}$, where $c_s \geq 1$ and $c_t \geq 1$ are constants, we consider $m=n$.
\end{itemize}

\caption{Asymmetric mutation}
\label{alg:asym}
\end{algorithm}

\begin{figure*}[t]
\centering
 
    \includegraphics[width=0.24\linewidth]{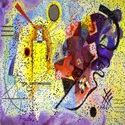}
    \includegraphics[width=0.24\linewidth]{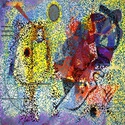}
    \includegraphics[width=0.24\linewidth]{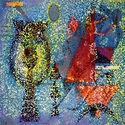}
    \includegraphics[width=0.24\linewidth]{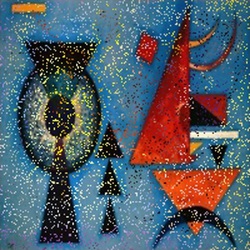}
\caption{Image Transition using asymmetric mutation with $c_s=100$ and $c_t=50$ at 12.5\%, 37.5\%, 62.5\% and 87.5\% of the target image (from left to right).}
\label{fig:3}
\end{figure*}

As the baseline mutation operator, we consider the asymmetric mutation operator which has been studied in the area of runtime analysis for special linear functions~\citep{DBLP:journals/ec/JansenS10} as well as the minimum spanning tree problems~\citep{NeumannW07}.
Using this mutation operator instead of standard bit mutations for \ONEMAX problems avoids the coupon collector's effect~\citep{Kobza2007,Mitz2005}. In the transition process, the coupon collector's effect implies that the last (even small) fraction of the pixels that need to be flipped to the target need more time to be flipped than all the pixels previously set to the target. More precisely, flipping each bit with probability $1/n$ as done in standard bit mutations, implies that the waiting time to flip the last pixel to the target is $\Theta(n)$ whereas increasing the number of target pixels is $\Theta(1)$ if the number of target pixels is still a constant fraction of all the pixels.
Such a slow-down at the end of the transition process where  there is no progress for many iterations is not desirable.

We use the generalization of this asymmetric mutation operator already proposed in~\cite{DBLP:conf/iconip/NeumannAN16} and shown in Algorithm~\ref{alg:asym}.
Let $|X|_T$ be the number of pixels where $X$ and $T$ agree. Similarly, let $|X|_S$ be the number of pixels where $X$ and $S$ agree. Each pixel in starting state $s$ is flipped with probability $c_s/(2|X|_S)$ and each pixel in target state $t$ is flipped with probability $c_t/(2|X|_T)$. The special case of $c_s=c_t=1$ has been mathematically analyzed with respect to the runtime behaviour on other pseudo-Boolean functions.

We set the parameters as follows: $c_s$ = 100 and $c_t$ = 50. This allows both a decent and sufficient speed for the image transition process and enough exchanges of pixels for an interesting evolutionary process. We should mention that obtaining the last pixels of the target image may take a long time compared to the other progress steps when using large values of $c_t$. However, for image transition, this only effects steps when there are at most $c_t/2$ source pixels remaining in the image. From a practical perspective, this means that the evolutionary process has almost converged towards the target image and setting the remaining missing target pixels to their target values provides an easy solution.

All experimental results for evolutionary image transition in this paper are shown for the process of moving from the starting image to the target image given in Figure~\ref{fig:2} where the images are of size $200 \times 200$ pixels. The algorithms have been implemented in MATLAB.
In order to visualize the process, we show the images obtained when the evolutionary process reaches $12.5\%$, $37.5\%$, $62.5\%$ and $87.5\%$ pixels of target image for the first time. We should mention that all processes except the use of the biased random walks are independent of the starting and target image which implies that the use of other starting and target images would show the same effects in respect to the way that target pixels are displayed during the transition process.

In Figure~\ref{fig:3} we show the experimental results of the asymmetric mutation approach as the baseline.
On the first image from left we can see the starting image $S$ with lightly stippling dots in randomly chosen areas of the target image $T$. Consequently, the area of the yellow dimensional abstract face disappears, and black and red abstract figure appears. Meanwhile the background has adopted a dot pattern, where a nuance of dark and light develops steadily. In the last image, we barely see the starting image $S$ and the target image $T$ appearing permanently with the background becoming a darker blue tone, whereby the stippling effect shown in the middle two frames decreases. Interesting images in respect to aesthetic and evolutionary creativity emerge for the pictures at $37.5\%$ and $62.5\%$ of the evolutionary processes. In the third picture, we can observe elements of both images compounded with a very special effect as a result of the image transition process.

\section{Random Walks for Image Transition}
\label{sec3}

\begin{algorithm}[t]
\vspace{0.1cm}
\begin{itemize}
\item Choose the starting pixel $X_{ij} \in X$ uniformly at random.
\item Set $X_{ij} := T_{ij}$.
\item while (not termination condition)
\begin{itemize}
\item Choose $X_{kl} \in N(X_{ij})$ uniformly at random.
\item Set $i:=k$, $j:=l$ and $X_{ij}:=T_{ij}$. 
\end{itemize}
\item Return $X$.
\end{itemize}

\caption{Uniform Random Walk}
\label{alg:walk}
\end{algorithm}

\begin{algorithm}[t]

\vspace{0.1cm}
\begin{itemize}
\item Choose the starting pixel $X_{ij} \in X$ uniformly at random.
\item Set $X_{ij} := T_{ij}$.
\item while (not termination condition)
\begin{itemize}
\item Choose $X_{kl} \in N(X_{ij})$ according to probabilities $p(X_{kl})$.
\item Set $i:=k$, $j:=l$ and $X_{ij}:=T_{ij}$. 
\end{itemize}
\item Return $X$.
\end{itemize}

\caption{Biased Random Walk}
\label{alg:walk2}
\end{algorithm}

\begin{figure*}[h]
\centering

    \includegraphics[width=0.24\linewidth]{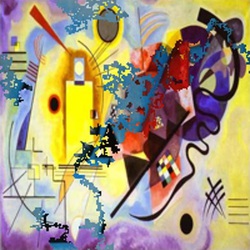}
    \includegraphics[width=0.24\linewidth]{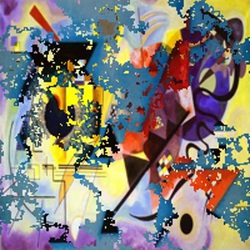}
    \includegraphics[width=0.24\linewidth]{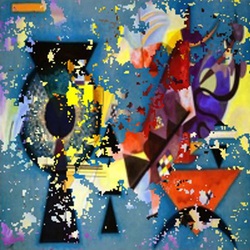}
    \includegraphics[width=0.24\linewidth]{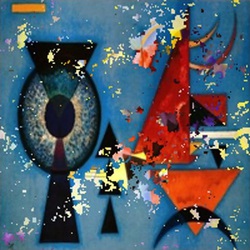}\\
  \vspace{0.3cm} 
       \includegraphics[width=0.24\linewidth]{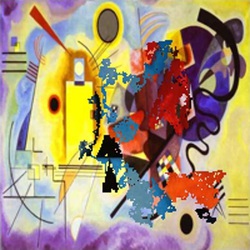}
       \includegraphics[width=0.24\linewidth]{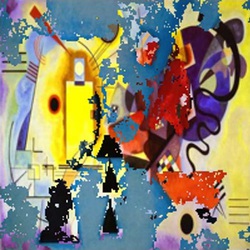}
       \includegraphics[width=0.24\linewidth]{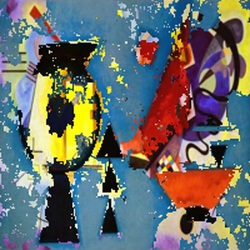}
       \includegraphics[width=0.24\linewidth]{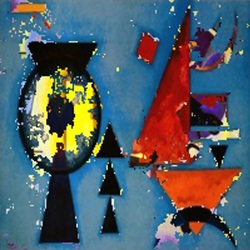}

\caption{Image Transition for Uniform Random Walk (top) and Biased Random Walk (bottom) with 12.5\%, 37.5\%, 62.5\% and 87.5\% of the target image (from left to right).}
\label{fig:4}
\end{figure*}

Our evolutionary algorithms for image transition build on random walk algorithms and use them later on as part of a mutation step.
Specifically, we investigate the use of random walk algorithms for image transition which move, at each step, from a current pixel $X_{ij}$ to one of 4-connected pixels in its neighbourhood. 

We utilize the neighbourhood $N(X_{ij})$ of $X_{ij}$ following von Neumann's definition of neighbourhood~\citep{Neumann:1966:TSA:1102024} as

\[
N(X_{ij}) = \{ X_{(i-1)j}, X_{(i+1)j}, X_{i(j-1)} X_{i(j+1)} \}
\]
where we work modulo the dimensions of the image in the case that the values leave the pixel ranges, $i \in \{1, \ldots, m\}$, $j \in \{1, \ldots, n\}$. This implies that from a current pixel, we can move up, down, left, or right. Furthermore, we wrap around when exceeding the boundary of the image. We do this in order to have a more interesting process. Furthermore, it supports the effectiveness of our biased random walks in the case that they would be biased towards the image boundary. Not wrapping around in this case would imply that pixels opposite of the focused boundary can hardly be reached.

The classical random walk shown in Algorithm~\ref{alg:walk} chooses an element $X_{kl} \in N(X_{ij})$ uniformly at random~\citep{pearson1905problem}. We call this the \emph{uniform random walk} in the following. The cover time of the uniform random walk on a $n \times n$ torus is upper bounded by $4n^2 (\log n)^2/\pi$~\citep{Dembo2004}  which implies that the expected number of steps of the uniform random walk until the target image is obtained (assuming $m=n$) is upper bounded by $4n^2 (\log n)^2/ \pi))$.

\subsection{Biased Random Walk}

We also consider a \emph{biased random walk} where the probability of choosing the element $X_{kl}$ is dependent on the difference in RGB-values for $T_{ij}$ and $T_{kl}$.
We favour a neighbor $X_{kl} \in N(X_{ij})$ if $T_{kl}$ is close to $T_{ij}$ in respect to RGB-values. The approach is shown in Algorithm~\ref{alg:walk2}.
Weighted random walks have been used in a similar way in the context of image segmentation~\citep{DBLP:journals/pami/Grady06}.
We denote by $T_{ij}^r$, $1 \leq r \leq 3$, the $r$th RGB value of $T_{ij}$ and define

\[
\gamma (X_{kl}) = \left(\max \left\{\sum_{r=1}^3 |T_{kl}^r - T_{ij}^r|,1 \right\}\right)^{\alpha},
\]
where $\alpha\geq 0$.


In our random walk, we prefer $X_{kl}$ if $\gamma (X_{kl})$ is small compared to the other elements in $N(X_{ij})$.
 In order to compute the probability of moving to a new neighbour we consider $(1/ \gamma( X_{kl})) \in [0,1]$ and prefer elements in $N(X_{ij})$ where this value is large.

In the biased random walk, the probability of moving from $X_{ij}$ to an element $X_{kl}  \in N(X_{ij})$ is given by

\begin{eqnarray*}
p(X_{kl}) &=& \frac{(1/\gamma(X_{kl}))}{\sum_{X_{st} \in N(X_{ij})} (1/\gamma(X_{st}))}.\\
\end{eqnarray*}

Our standard biased random walk works with $\alpha=1$ and we extend our investigations to other choices of $\alpha$ as part of our random walk mutation operator in Section~4.3. Furthermore, we use different choices of $\alpha$ for image painting in Section~7. It should be noted that the uniform random walk given in Algorithm~\ref{alg:walk} is the special case of Algorithm~\ref{alg:walk2} where $\alpha=0$.

The biased random walk is dependent on the target image when carrying out mutation or random walk steps and the importance of moving to a pixel with similar color.
By introducing the bias in respect to pixels that are similar, the bias can cause the evolutionary image transition process to take exponentially long as the walk might encounter effects similar to the gambler's ruin process~\citep{Mitz2005}. For our combined approaches described in the next section, we use the random walks as mutation components which ensures that the evolutionary image transition is carried out efficiently. We will use the biased random walk for evolutionary image transition in Section~\ref{sec4}.

In Figure~\ref{fig:4} we show the experimental results of the uniform random walk and biased random walk.
At the beginning, we can observe the image with the characteristic random walk pathway appearing as a patch in the starting image $S$. Through the transition process, the clearly recognisable patches on the target image $T$ emerge. In the advanced stages, the darker patches from the background of the target image dominate. The effect in animation is that the source image is scratched away in a random fashion to reveal an underlying target image.

The four images of the biased random walk are clearly different to the images of the uniform random walk. During the course of the transition, the difference between these processes becomes more prominent, especially in the background where at $87.5\%$ pixels of the target image there is nearly an absolute transition to the target image $T$. In strong contrast, the darker abstract figure of the images stays nearly untouched, so that we see a layer of the yellow face in starting image $S$ in the center of the abstract black figure in target image $T$. In this image, the figure itself is also very incomplete with much of the source picture showing through. These effects arise from biased probabilities in the random walk which makes it difficult for the walk to penetrate areas of high contrast to the current pixel location. 

\section{Random Walk Mutation and Combined Approaches}
\label{sec4}

The asymmetric mutation operator and the random walk algorithms display quite different behaviour when applied to image transition. We now study additional ways of carrying out mutations as part of the image transition process. Our goal is to obtain a new evolutionary process using mutations based on random walks and biased random walks. Furthermore, we investigate the effect of combining them with the asymmetric mutation operator.

\subsection{Random Walk Mutation}
\label{sec:rwmutation}

\begin{figure*}[t]
\centering

    \includegraphics[width=0.24\linewidth]{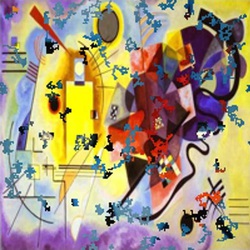}
    \includegraphics[width=0.24\linewidth]{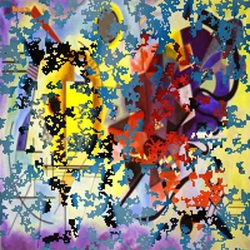}
    \includegraphics[width=0.24\linewidth]{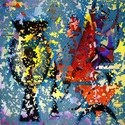}
    \includegraphics[width=0.24\linewidth]{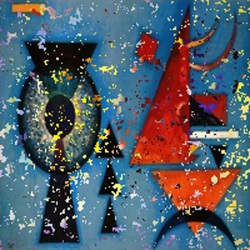}\\

    \vspace{0.3cm}    
       \includegraphics[width=0.24\linewidth]{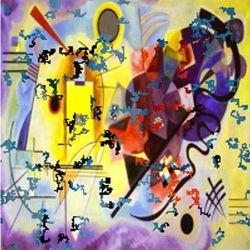}
       \includegraphics[width=0.24\linewidth]{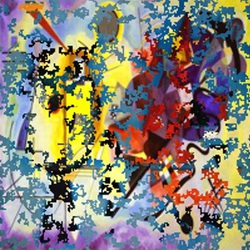}
       \includegraphics[width=0.24\linewidth]{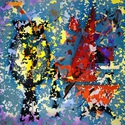}
       \includegraphics[width=0.24\linewidth]{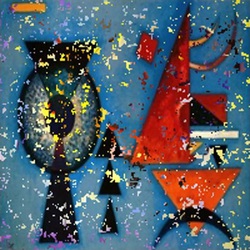}

\caption{Image Transition for \rwalk (top) and \bwalk (bottom) with 12.5\%, 37.5\%, 62.5\% and 87.5\% of the target image (from left to right).}
\label{fig:5X}
\end{figure*}

Firstly, we explore the use of random walks as mutation operators and call this a \emph{random walk mutation}.

The \emph{uniform random walk mutation} selects the position of a pixel $X_{ij}$ uniformly at random and runs the uniform random walk for  $\tmax$ steps (iterations of the while-loop). We call the resulting algorithm \rwalk.
Similarly, the \emph{biased random walk mutation} selects the position of a pixel $X_{ij}$ uniformly at random and runs the biased random walk for $\tmax$ steps. This algorithm is called \bwalk. 
For our experiments, we set $\tmax=100$ which implies that each mutation carries out a random walk consisting of $100$ steps.

Figure~\ref{fig:5X} shows the results of the experiments for \rwalk and \bwalk. The transitions produce significantly different images compared to the previous ones. In both experiments we can see the target image emerging through a series of small patches at first, then steadily changing through a more chaotic phase where elements of the source and target image appear with roughly equal frequency. The last image of each experiment is most similar to the target image.

The images from \bwalk appear visually similar to those of the \rwalk in the beginning but differences emerge during the final stages of the transition. In \bwalk,
elements of the source image still show through in areas of high contrast in the target image, which the biased random walk finds difficult to traverse. At a more local scale, this mirrors the effects of bias in the earlier random walk experiments. At a global scale, it can be seen that the blue background, which is a low contrast area, is slightly more complete in the final frame of \bwalk than in the same frame in \rwalk.

\subsection{Impact of Random Walk Length} 
\label{sec4new}
%
Furthermore, we investigate the impact of the random walk length in \brw mutation. We want to explore how different lengths of \brw influence the creative process of evolving images. We also aim to understand the influence that the length of the \brw has on the results.

In general, we want to investigate the importance of different random walk lengths in the evolutionary process. Also want to understand how this parameter shapes the final result and how we are able to control those effects in a systematically manner. The artistic techniques that appear from evolutionary image transition using different random walks can be compared to a French style of abstract painting, namely Tachisme (synonymous with art informel). Tachisme features the intuitive and spontaneous gestures of the artist's brushstrokes that involve the use of dabs, drips, or splotches of colour.

\begin{figure*}[!ht]
\centering
 
    \includegraphics[width=0.24\linewidth]{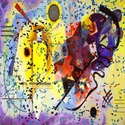}
    \includegraphics[width=0.24\linewidth]{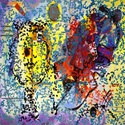}
    \includegraphics[width=0.24\linewidth]{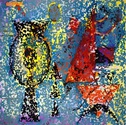}
    \includegraphics[width=0.24\linewidth]{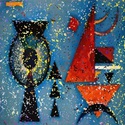}\\
  
 \vspace{0.3cm} 
 
    \includegraphics[width=0.24\linewidth]{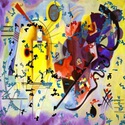}
    \includegraphics[width=0.24\linewidth]{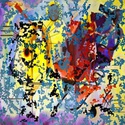}
    \includegraphics[width=0.24\linewidth]{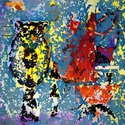}
    \includegraphics[width=0.24\linewidth]{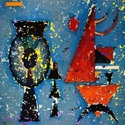}\\

    \vspace{0.3cm} 
    
       \includegraphics[width=0.24\linewidth]{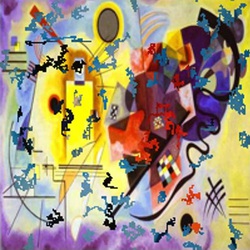}
       \includegraphics[width=0.24\linewidth]{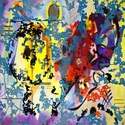}
       \includegraphics[width=0.24\linewidth]{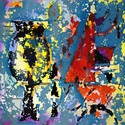}
       \includegraphics[width=0.24\linewidth]{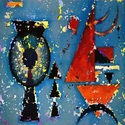}

    \vspace{0.3cm}    
       \includegraphics[width=0.24\linewidth]{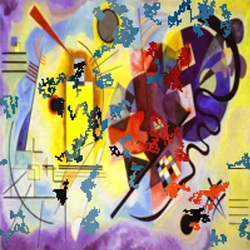}
       \includegraphics[width=0.24\linewidth]{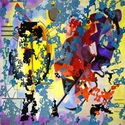}
       \includegraphics[width=0.24\linewidth]{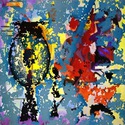}
       \includegraphics[width=0.24\linewidth]{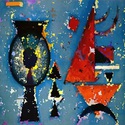}
       

\caption{Image Transition for \brw with $t_{\max}= 10, 50, 200, 400$ (from top to bottom) and 12.5\%, 37.5\%, 62.5\% and 87.5\% of the target image (from left to right). }
\label{fig:5Xdiftmax}
\end{figure*}

In our previous experiments described in Section \ref{sec:rwmutation}, we used the \rw for evolutionary image transition and set the length of the \rw and the \brw to $100$. Now, we investigate different choices of $\tmax$: small ($\tmax = 10, 50$), middle ($\tmax = 200, 400$) and large ($\tmax = 2000, 10000, 20000, 50000$). In Figure \ref{fig:5Xdiftmax} we present the experimental results.
In the case of using different length of \rw, the special characteristics of \brw play the most important role as the \brw has tendencies of moving to similar colors.

At the beginning, we observe that the use of parameter $\tmax = 10$ and $\tmax = 50$ results in images with smaller patches.
In the later stages, the darker patches from the background dominate the image. Apparently differences between image transition with small and middle $\tmax$ occur. It can be observed that the small and middle random walk lengths produce various distinctive images.

When the process of the evolutionary image transition reaches $12.5\%$ and $37.5\%$ pixels of target image, we can observe more chaotic \rw patches that are present over image with tendency to an isolated structure. 
At the same stage of the evolutionary image processes, the images produced with middle $\tmax$ are less patchy and have a less irregular appearance. The reason for this is that the \brw build longer pads during the evolutionary transition.

Furthermore, we notice differences in the images during the evolutionary processes when $62.5\%$ and $87.5\%$ pixels of target image occur. Here, the image transition for \brw with middle $\tmax$ has more patches incorporated into the target image $T$. Thus, the images displayed obtain a nearly chaotic state. In the examples described above, we can see that the different lengths of the \brw have a clear effect on the image transition process.

\begin{figure*}[!t]
\centering

    \vspace{0.3cm}    
       \includegraphics[width=0.24\linewidth]{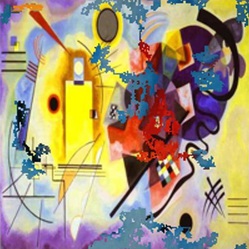}
       \includegraphics[width=0.24\linewidth]{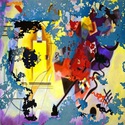}
       \includegraphics[width=0.24\linewidth]{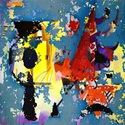}
       \includegraphics[width=0.24\linewidth]{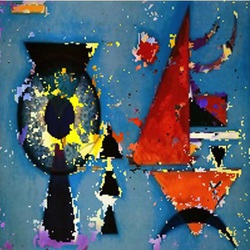}

    \vspace{0.3cm}    
       \includegraphics[width=0.24\linewidth]{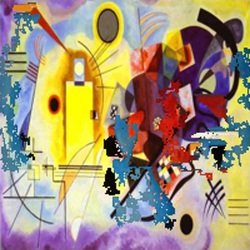}
       \includegraphics[width=0.24\linewidth]{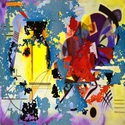}
       \includegraphics[width=0.24\linewidth]{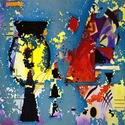}
       \includegraphics[width=0.24\linewidth]{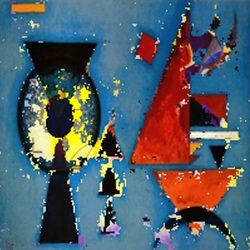}
       
        \vspace{0.3cm}    
       \includegraphics[width=0.24\linewidth]{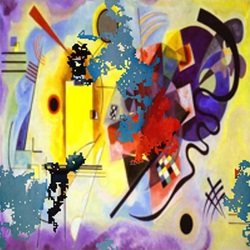}
       \includegraphics[width=0.24\linewidth]{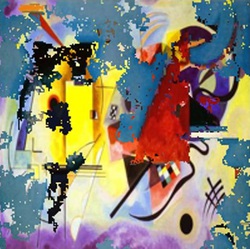}
       \includegraphics[width=0.24\linewidth]{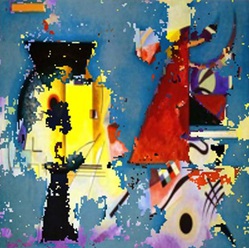}
       \includegraphics[width=0.24\linewidth]{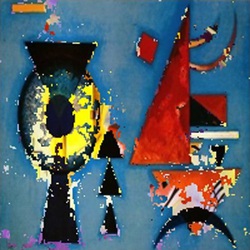}
       
              \vspace{0.3cm}    
       \includegraphics[width=0.24\linewidth]{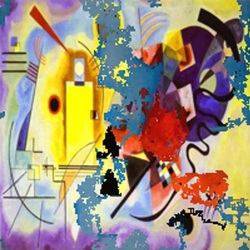}
       \includegraphics[width=0.24\linewidth]{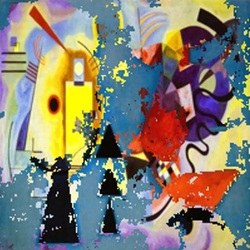}
       \includegraphics[width=0.24\linewidth]{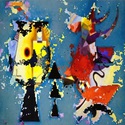}
       \includegraphics[width=0.24\linewidth]{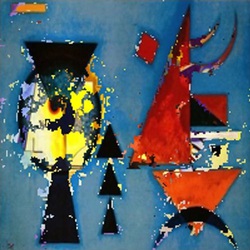}
       
       \caption{Image Transition for \bwalk with $t_{\max}=2000, 10000, 20000, 50000$ (from top to bottom) and 12.5\%, 37.5\%, 62.5\% and 87.5\% of the target image (from left to right).}
\label{fig:5Xdiftmax2}
\end{figure*}

In Figure \ref{fig:5Xdiftmax2} we present the experimental results of the \brw with length of $2000$, $10000$, $20000$ and $50000$, respectively. We can observe that the resulting images obtained with larger $\tmax$ are visual different from the images obtained with smaller length of \rw.

At the beginning, when the process of the evolutionary image transition reaches $12.5\%$ and $37.5\%$ pixels of target image, we can observe only a few larger patches of the target image. As consequence of that, we see start image $S$ dominates during the evolutionary image transition.
In the later stage, when $62.5\%$ and $87.5\%$ pixels of target image occur, we see that the darker abstract figure is overall untouched. This is the result of the bias occurring during the transition with \brw mutation. In contrast to the experiments conducted with small and medium $\tmax$, the images are mostly completed with more recognizable elements from starting $S$ and target $T$ image. Using \brw with a larger $\tmax$ into evolutionary image transition can create images that maintain content of both images without losing the style of the images.

\subsection{Impact of the choice of \textbf{$\alpha$} in the Biased Random Walk} 
\label{sec5new}

 \begin{figure*}[t!]
\centering

    \includegraphics[width=0.24\linewidth]{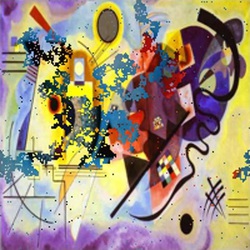} 
    \includegraphics[width=0.24\linewidth]{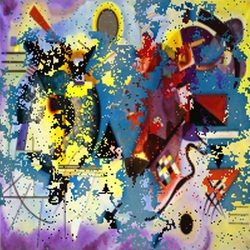}
    \includegraphics[width=0.24\linewidth]{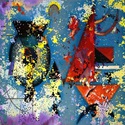}
    \includegraphics[width=0.24\linewidth]{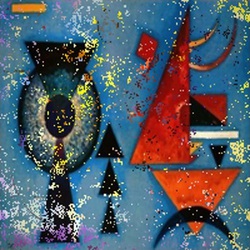}\\
      \vspace{0.3cm}

         \includegraphics[width=0.24\linewidth]{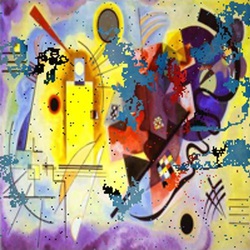}  
       \includegraphics[width=0.24\linewidth]{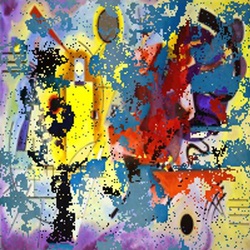}
      \includegraphics[width=0.24\linewidth]{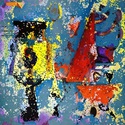}
     \includegraphics[width=0.24\linewidth]{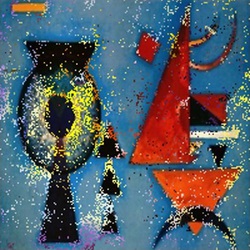}

\caption{Image Transition for \arwalk (top) and \abwalk (bottom) with 12.5\%, 37.5\%, 62.5\% and 87.5\% of the target image (from left to right).}
\label{fig:8}

\end{figure*}

Additionally, we investigate the impact of the choice of the $\alpha$ in \brw mutation. 
We want to examine how the different $\alpha$ of \brw influence the generation of artistic images.
We aim to understand the influence that the choice of different $\alpha$ has on the results. 

In our previous experiments described in Section \ref{sec2}, we used the \brw for evolutionary image transition and set the $\alpha$ to $0$.

From now on, we investigate different choices of $\alpha$: small values ($\alpha = 0.25, 0.50$), middle values ($\alpha = 0.75, 1.0, 1.25$) and large values ($\alpha = 1.5, 2.0$). In Figure \ref{fig:7_2alphaEIT_BRW}, we observe the experimental results of our investigation.

The most differences can be observed in images during the evolutionary processes when $62.5\%$ pixels of target image occur. The evolutionary image transition for our algorithm with smaller value of $\alpha$ imply to have more patches incorporated into target image $T$. Thus, the image occurs more finished. In the examples described above, we can assume that the different values of $\alpha$ has predominantly influence in rather smaller degree with the chosen parameter $\tmax = 2000$, during the final stage of the evolutionary image transition algorithm.

\subsection{Combination of asymmetric and random walk mutation}
Furthermore, we explore the combination of the asymmetric mutation operator and random walk mutation.
Here, we run the asymmetric mutation operator as described in Algorithm~\ref{alg:asym} and a random walk mutation every $\tau$ generations. We explore two combinations, namely the combination of the asymmetric mutation operator with the uniform random walk mutation (leading to the algorithm \arwalk)
 as well as the combination of the asymmetric mutation operator with the biased random walk mutation (leading to Algorithm \abwalk). We set $\tau=1$ and $\tmax=2000$ which means that the process is alternating between asymmetric mutation and random walk mutation where each random walk mutation carries out $2000$ steps.

In Figure~\ref{fig:8}, we show the results of \arwalk and \abwalk. From a visual perspective both experiments combine the stippled effect of the asymmetric mutation
 with the patches of the random walk. In \abwalk there is a lower tendency for patches generated by random walks to deviate into areas of high contrast. As the experiment progresses, the pixel transitions caused by the asymmetric mutation have a tendency to degrade contrast barriers. 

However, even in the final frames there is clearly more background from the target image in \abwalk than in \arwalk. Moreover, there are more remaining patches of the source image near the edges of the base of  ct figure, creating interesting effects.

As it can be observed through the experimental results, the user obtains a large variety of effects for image transition through the choice of the different parameters. Large random walk lengths in the mutation operator lead to large patches in the transition process whereas small values of $\tmax$ imply that the are many smaller patches widely distributed across the image. This implies that the image transitions appears more evenly across the whole image. The combination with asymmetric mutation which flips single pixels allows for a further smoothening of the image transition process.

\section{Feature-based Analysis}
\label{sec6}

\begin{figure}[!th] 
\vspace{0.1cm}
\centering
\vskip -6pt
\vspace{0.1cm}
\includegraphics[scale=0.1477]{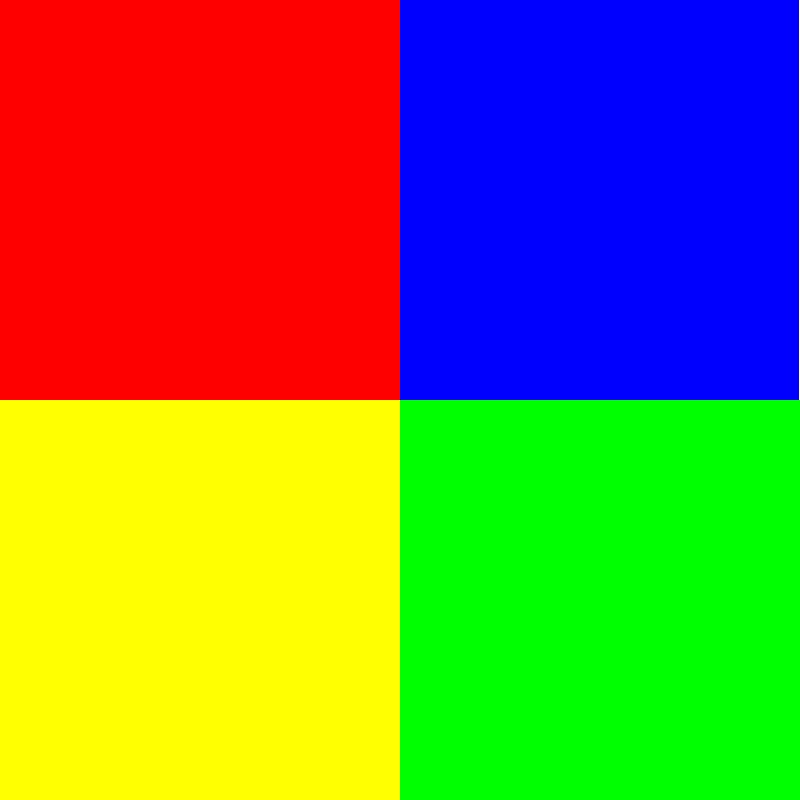}
\includegraphics[scale=0.1477]{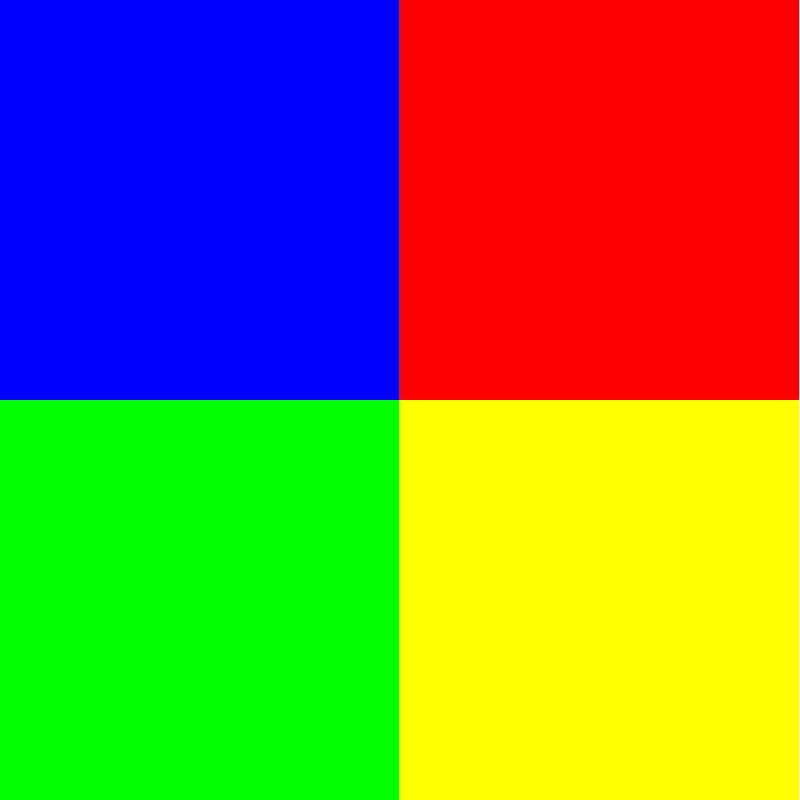}\\
\vspace{0.1cm}
\caption{Starting image $S$ (Color1) and target image $T$ (Color2).}
\label{fig:color}
\end{figure}

We now analyze the different introduced approaches for evolutionary image transition with respect to some features that measure aesthetic behaviour. Our goal is twofold. First, we analyze how the aesthetic feature values change during the process of transition. Furthermore, we compare the different approaches against each other and show where they differ with respect to the examined features when used for evolutionary image transition. For our investigations, we examine the starting and target image of Figure~\ref{fig:2}, the transition of a black starting image into a white target image, and the transition of the starting image Color1 into the target image Color2 as shown in Figure~\ref{fig:color}. Taking the last two pairs of images allows us to get additional systematic insights into the process of evolutionary image transition. Note that the images of Figure~\ref{fig:color} are only swapping the colored squares.

The set of features we use are, in order of appearance, \emph{Benford's Law}~\citep{jolion2001images}, \emph{Global Contrast Factor}~\citep{matkovic2005global}, \emph{Hue}, and \emph{Colorfulness}~\citep{hasler2003measuring}. We describe each of them in the following.

The {\emph{Benford's Law}} feature ($\Ben$) is a measure  of naturalness in an image $X$. 
Jolion~\citep{jolion2001images} observed that the sorted histogram of luminosities in natural images followed the 
shape of Benford's Law distribution of first digits. Here we use the encoding of the Benford's Law feature based on the one used by den Heijer~\citep{DBLP:journals/swevo/HeijerE14}. 

To calculate $\Ben(X)$ we first calculate a nine-bin histogram $H_X$ of the luminosities of $X$. The bins of $H_X$ are then sorted by frequency and scaled to sum to $1.0$. We define 
$$\Ben(X)=1- d_{total}/d_{max} $$
 where 
$$d_{total}=\sum_{i=1}^9 H_X(i)- H_{benford}(i)$$
and $H_{benford}$ is a 9-bin histogram, encoding Benford's Law distribution, with the bin frequencies $0.301, 0.176, 0.125, 0.097, 0.079, 0.067, 0.058, 0.051, 0.046$. 
Following \cite{DBLP:journals/swevo/HeijerE14}, we use
$$d_{max}=(1-H_{benford}(1))+\sum_{i=2}^9 H_{benford}(i)= 2 \cdot(1-H_{benford}(1))$$
which is the maximum deviation obtained if everything is assigned to bin $1$ under the above assumption that bins are sorted in decreasing order.

{\emph{Global Contrast Factor}}, $\GCF$ is a measure of mean contrast between neighbouring pixels at different image resolutions. To calculate $\GCF(X)$ we calculate the local contrast at each pixel at a given resolution $r$:  $\lc_r(X_{ij})=\sum_{X_{kl} \in N(X_{ij})} |\lum (X_{kl}) -\lum(X_{ij})|$ where $\lum(P)$ is the perceptual luminosity of pixel $P$ and $N(X_{ij})$ are the four neighbouring pixels of $X_{ij}$ at resolution $r$. The mean local contrast at the current resolution is defined $C_r=(\sum_{i=1}^m \sum_{j=1}^n \lc_r(X_{ij}))/(mn)$. From these local contrasts, $\GCF$ is calculated as
 $\GCF = \sum_{r=1}^9 w_r \cdot C_r.$

The pixel resolutions correspond to different {\em{superpixel}} sizes of $1,2,4,8,16,25,50,100$, and $200$. Each superpixel is set to the average luminosity of the pixel's it contains. The $w_r$ are empirically derived weights of resolutions from~\cite{matkovic2005global} giving highest weight to moderate resolutions.

The {\emph{Hue} of an image $X$ is 
$$\Hue(X) = \left(\sum_{i=1}^m \sum_{j=1}^n \mbox{\em h}(X_{ij})\right)/(m\times n)$$ where $\mbox{\em h}(X_{ij})$ is the hue value for pixel $X_{ij}$ in the range $[0,1]$. The function $\Hue$ measures where on average the image $X$ sits on the color spectrum. Because the color spectrum is a circular construct, one color, red in our case, is mapped to both 1 and 0. 

{\emph{Colorfulness}} ($\Col$) is a measure of the perceived variety of color in of an image. We use Hasler's simplified metric for calculating colorfulness~\citep{hasler2003measuring}. This measure quantifies spreads and intensities of opponent colors by calculating for the RGB values in each pixel $X_{ij}$ the red-green difference: $\mbox{\em{rg}}_{ij} =|R_{ij}-G_{ij}|$, and the yellow-blue difference: 
$\mbox{\em yb}_{ij}=|(R_{ij}+G_{ij})/2-B_{ij}|$. The means: $\mu_{rg}$, $\mu_{yb}$ and standard-deviations: $\sigma_{rg}$, $\sigma_{yb}$ for these differences are then combined to form a weighted magnitude estimate for colorfulness for the whole image:
\[
\Col(X)=\sqrt{\sigma_{rg}^2 + \sigma_{yb}^2} + 0.3 \sqrt{\mu_{rg}^2 + \mu_{yb}^2}
\]

 \begin{figure*}[htpb]
          \hfill  \hspace{-0.7cm}
          \subfigure[Benford's Law]{
          \includegraphics[width=0.4\linewidth]{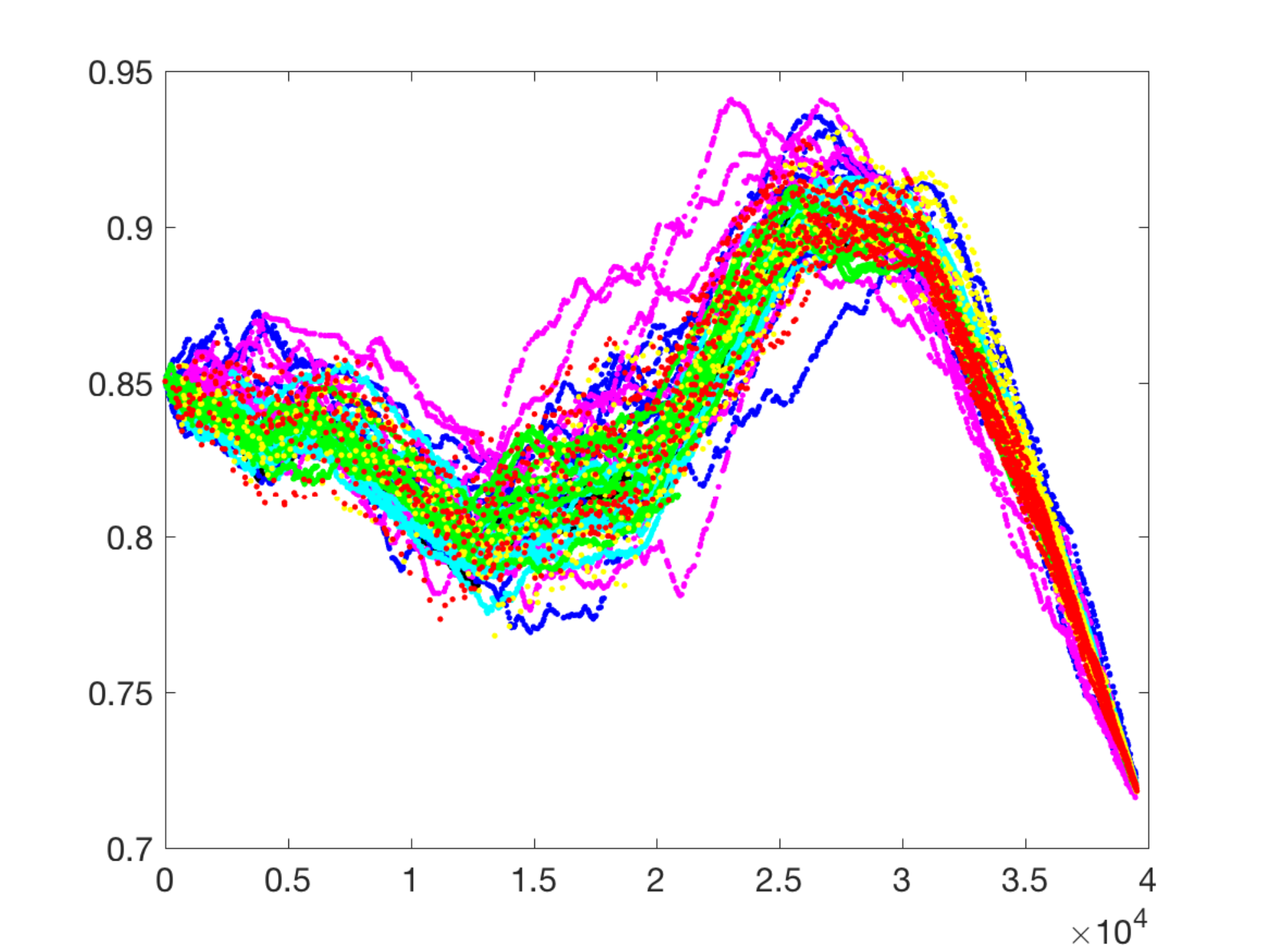}  \hspace{-0.6cm}
          \includegraphics[width=0.4\linewidth]{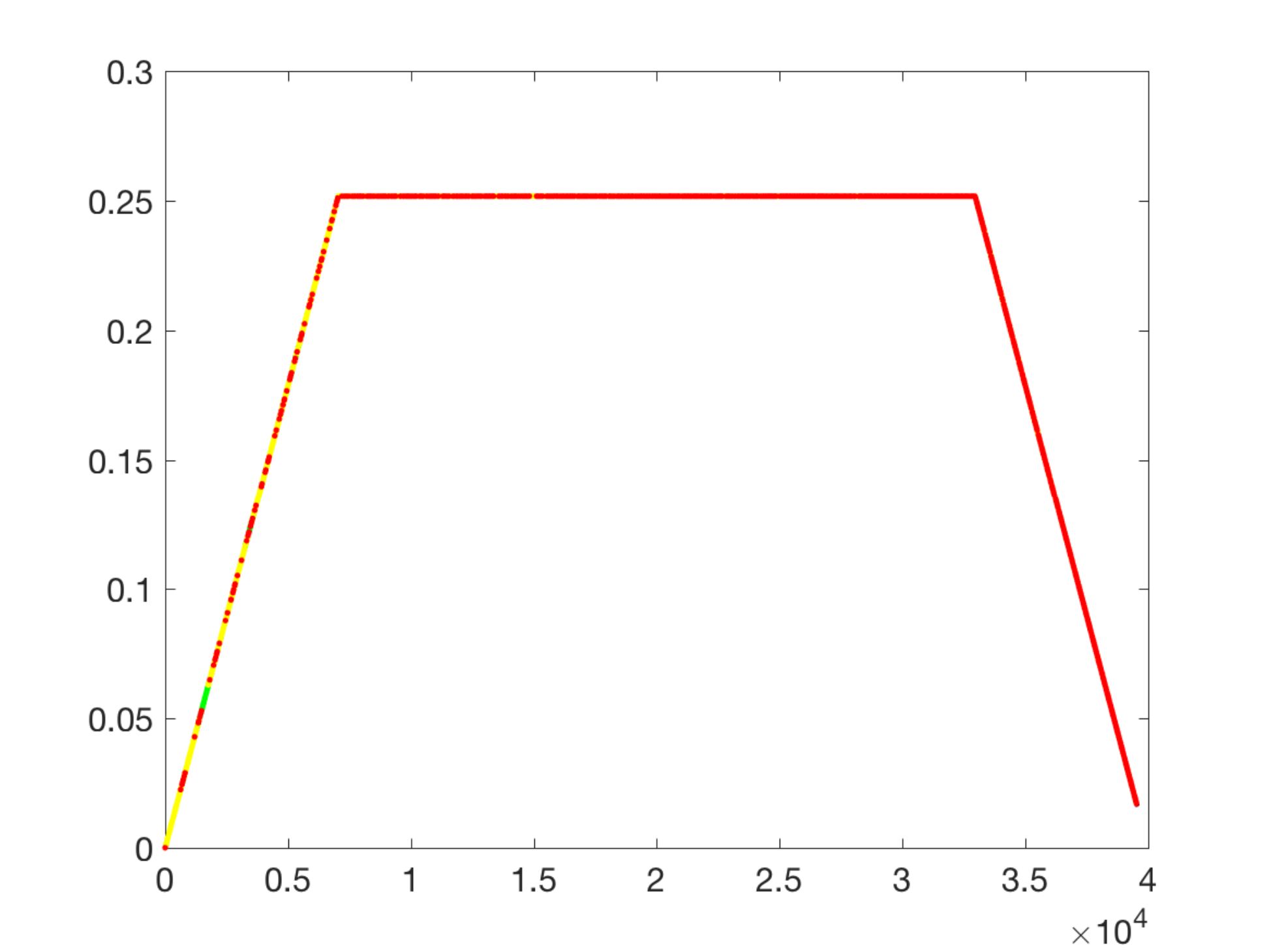}    \hspace{-0.6cm} 
           \includegraphics[width=0.4\linewidth]{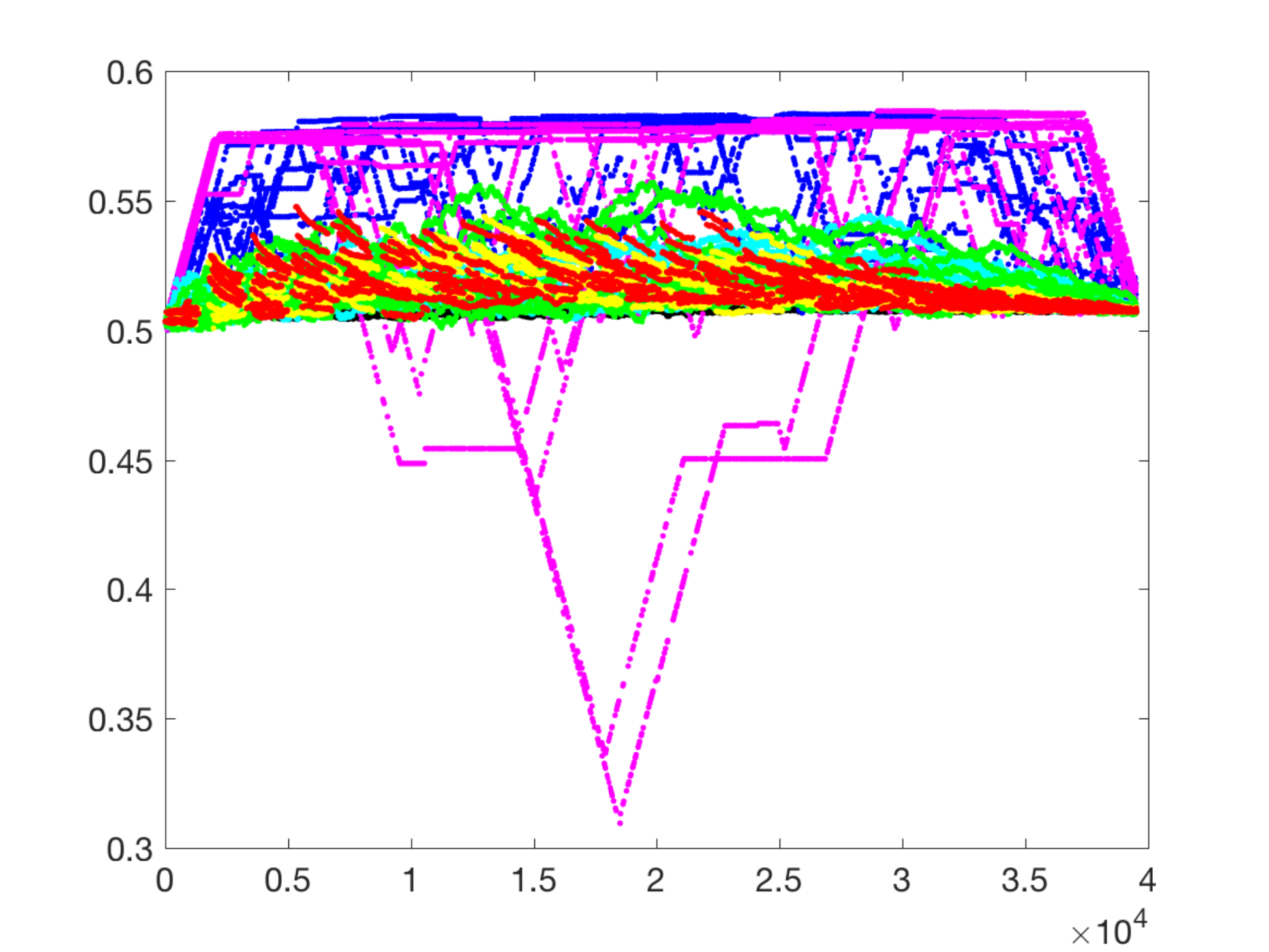}  \hspace{-0.6cm}
         
       }   
            \vspace{-0.2cm}
             \hfill  \hspace{-0.7cm}
             \subfigure[Global Contrast Factor]{
            \includegraphics[width=0.4\linewidth]{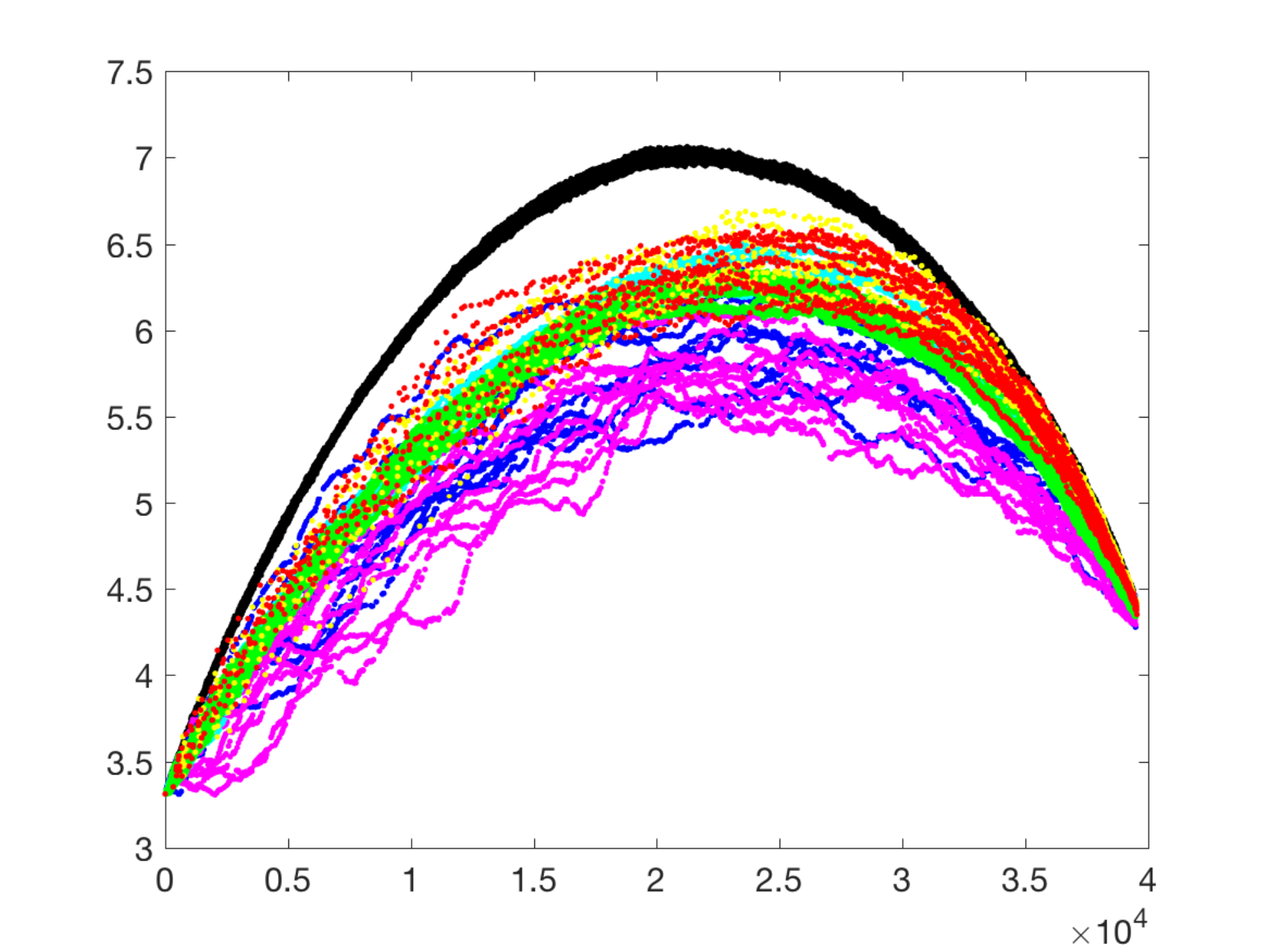} \hspace{-0.6cm}
            \includegraphics[width=0.4\linewidth]{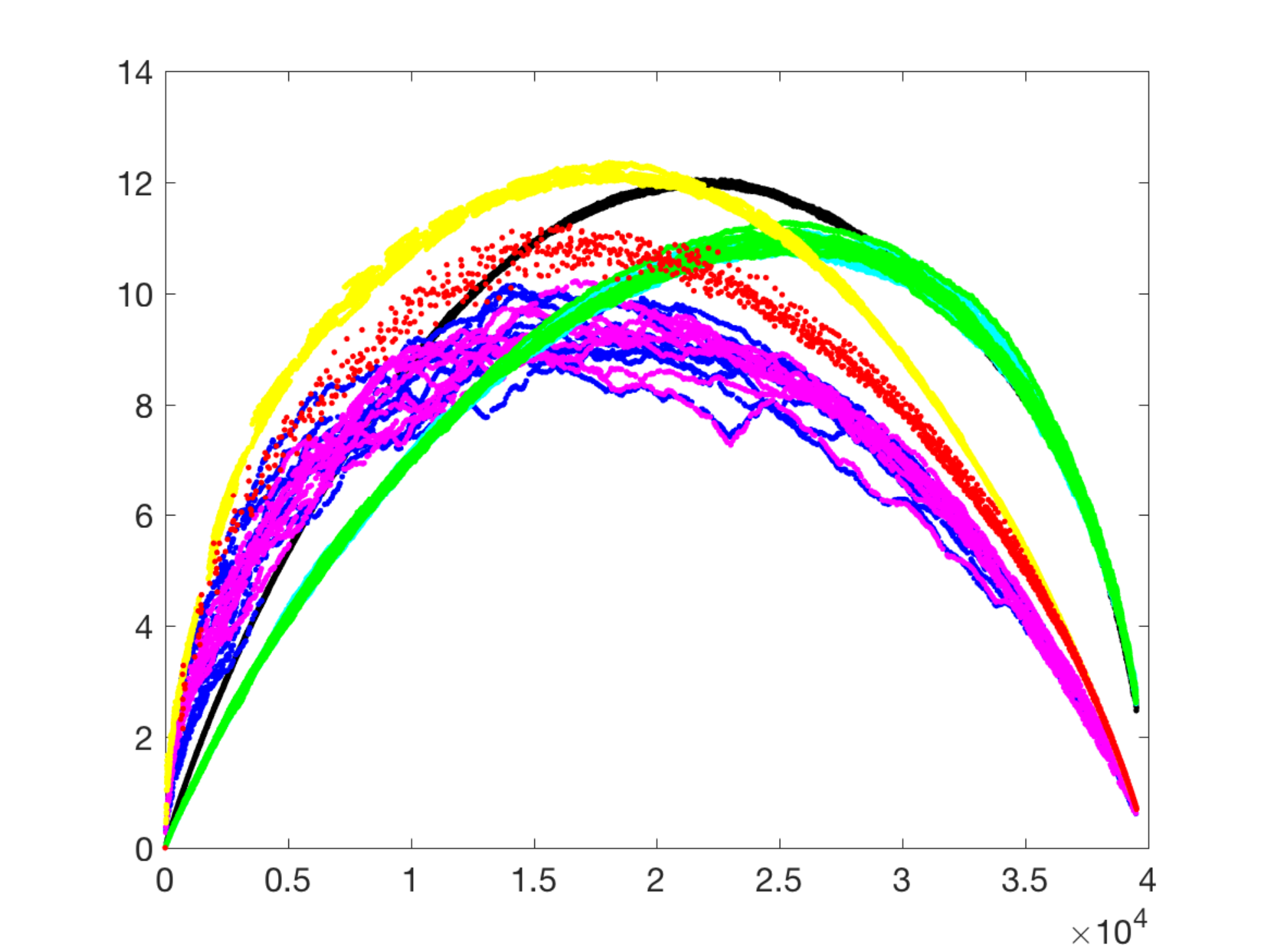} \hspace{-0.6cm}
            \includegraphics[width=0.4\linewidth]{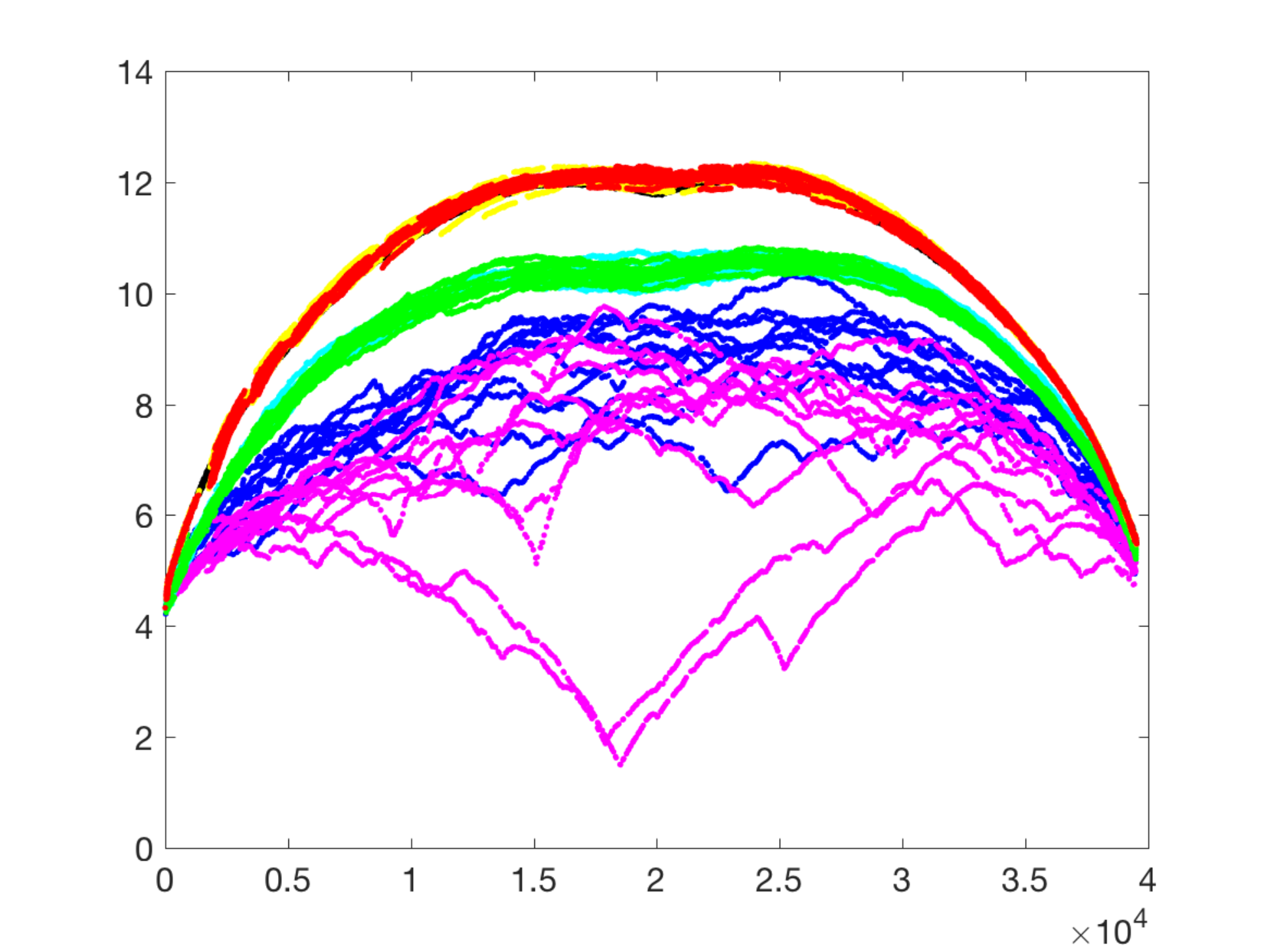} \hspace{-0.6cm}
           }
             \vspace{-0.2cm}
            \hfill  \hspace{-0.7cm}
            \subfigure[Hue]{
            \includegraphics[width=0.4\linewidth]{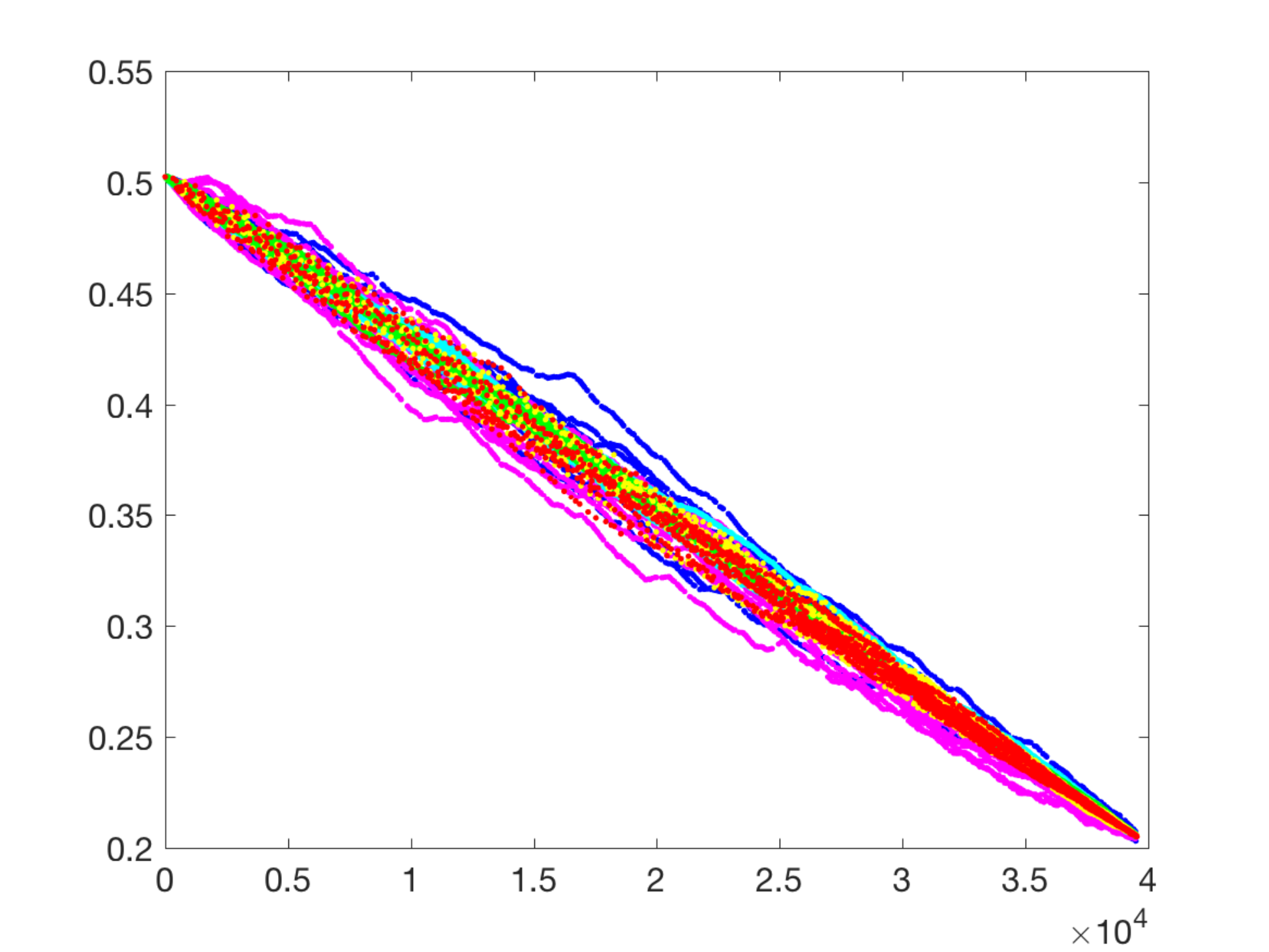} \hspace{-0.6cm}
            \includegraphics[width=0.4\linewidth]{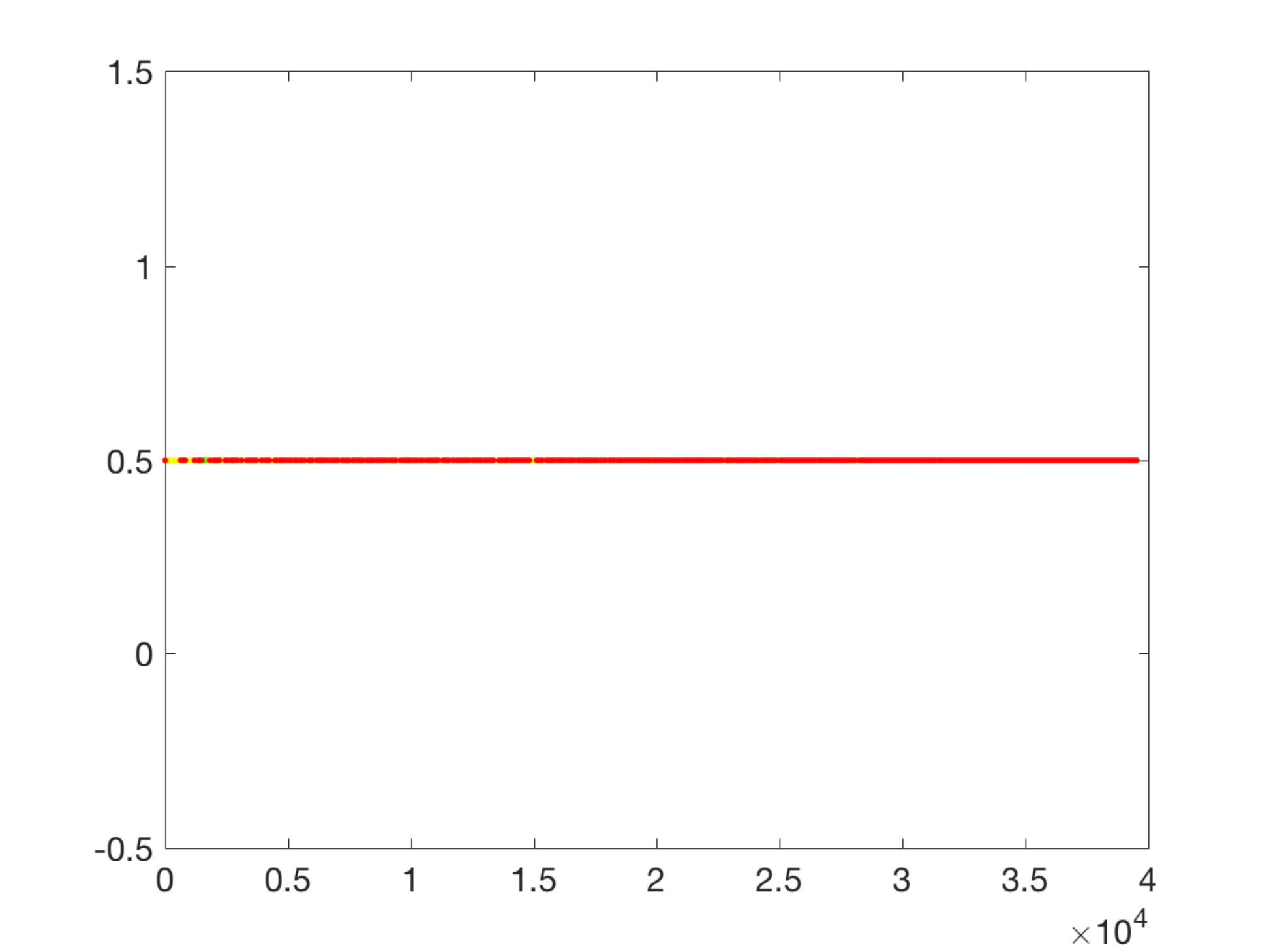} \hspace{-0.6cm}
            \includegraphics[width=0.4\linewidth]{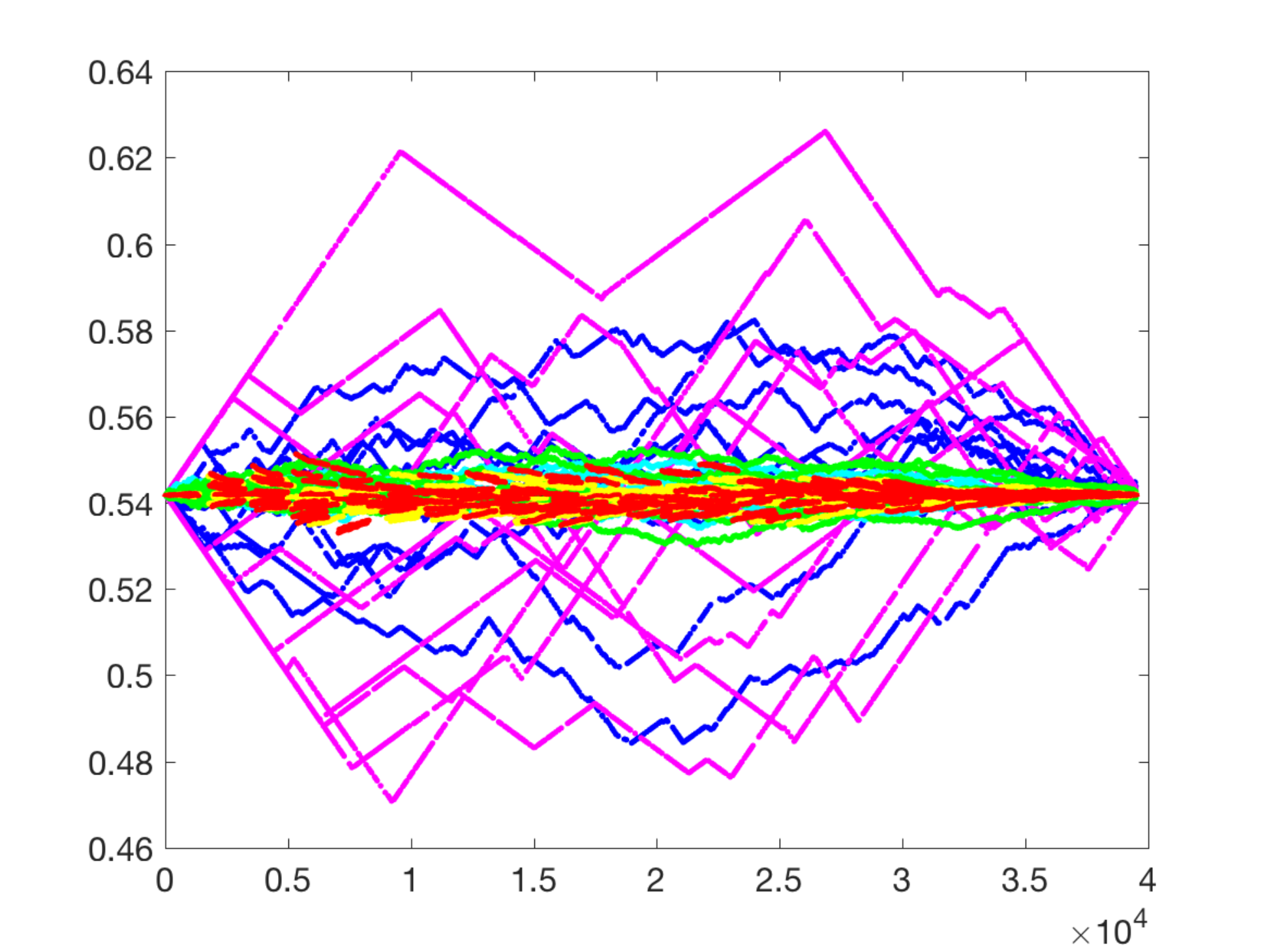} \hspace{-0.6cm}
            } 
            \vspace{-0.2cm}
             \hfill  \hspace{-0.7cm}
             \subfigure[Colorfulness]{
             \includegraphics[width=0.4\linewidth]{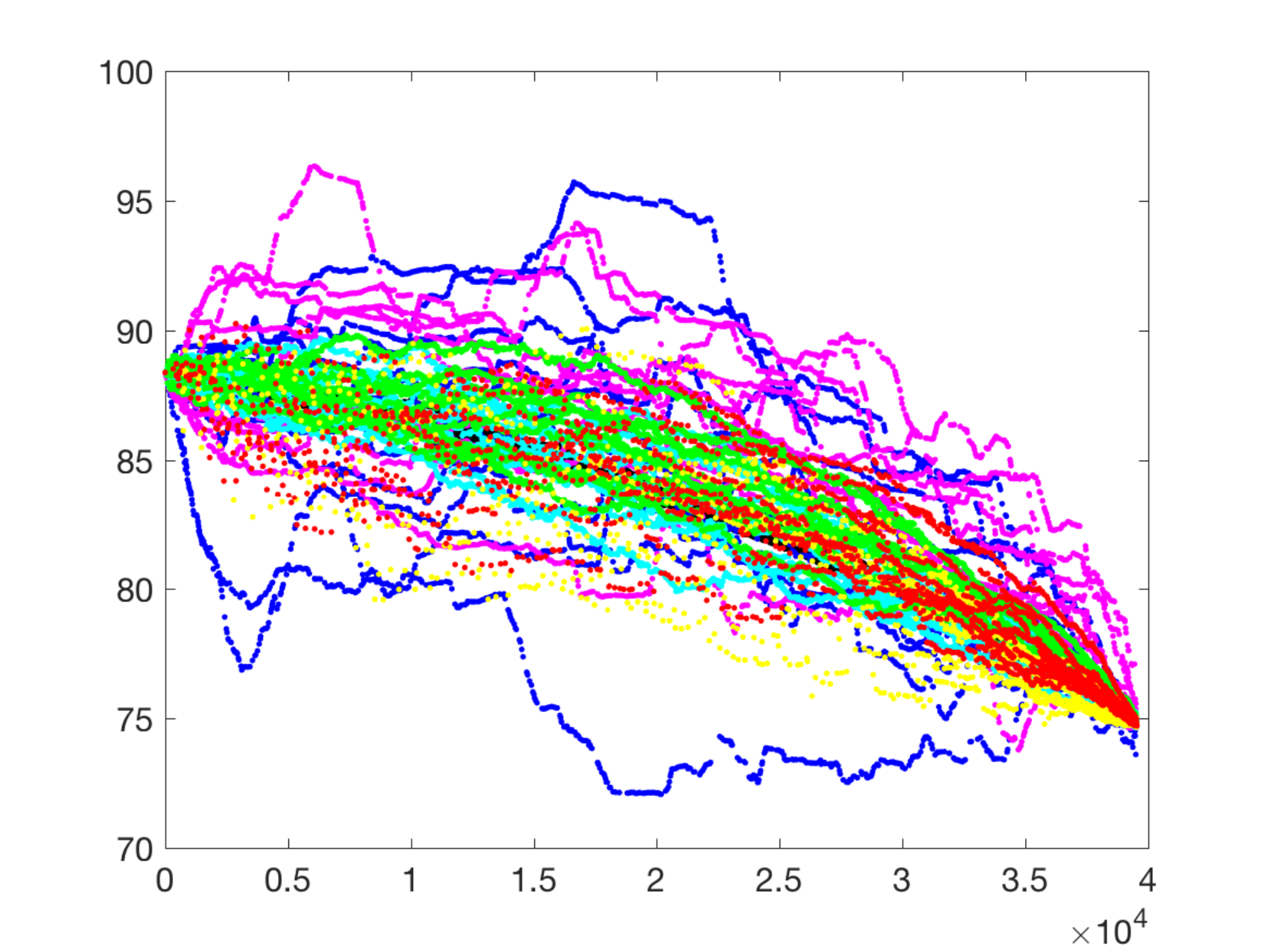} \hspace{-0.6cm}
             \includegraphics[width=0.4\linewidth]{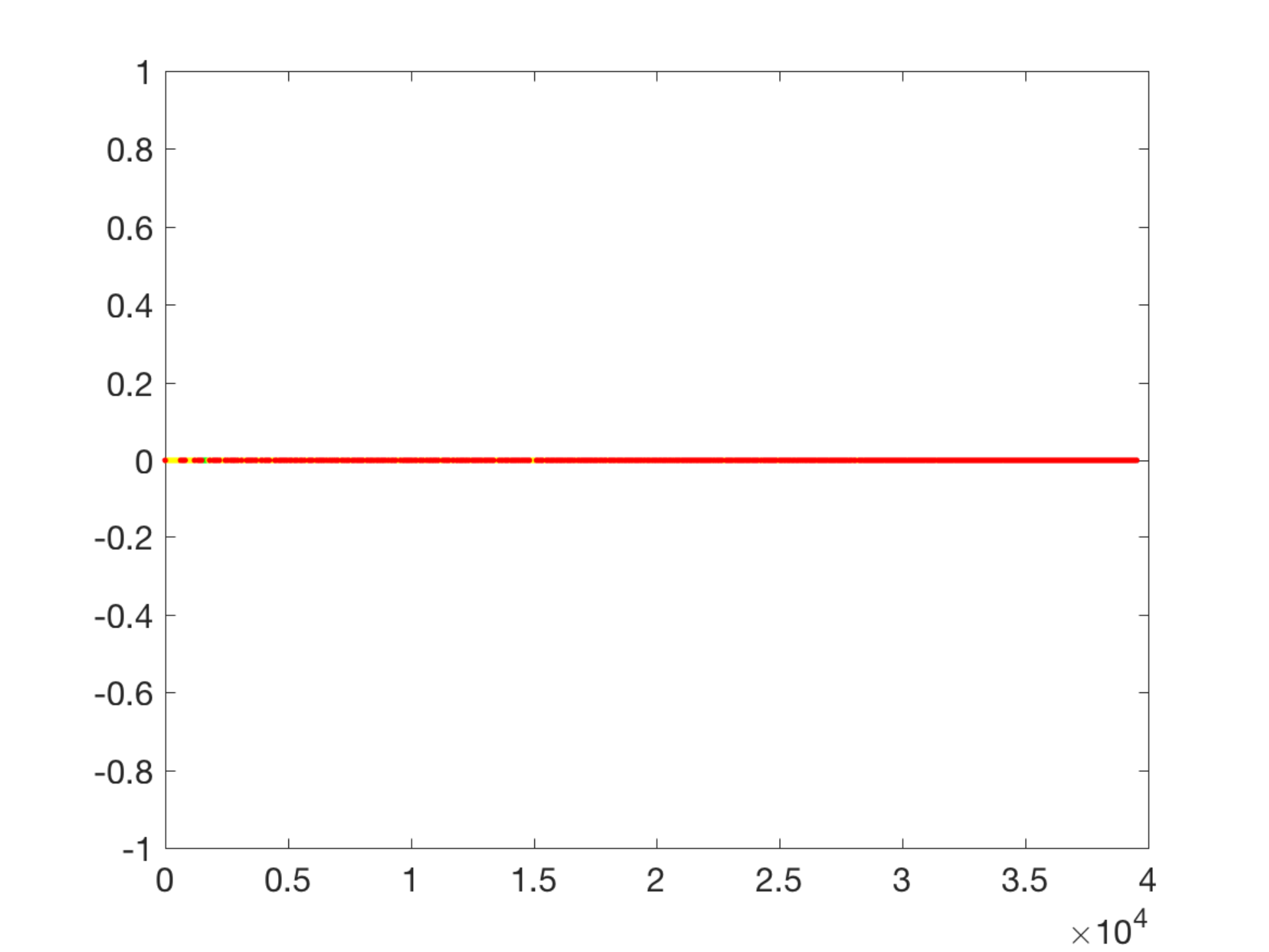} \hspace{-0.6cm}
              \includegraphics[width=0.4\linewidth]{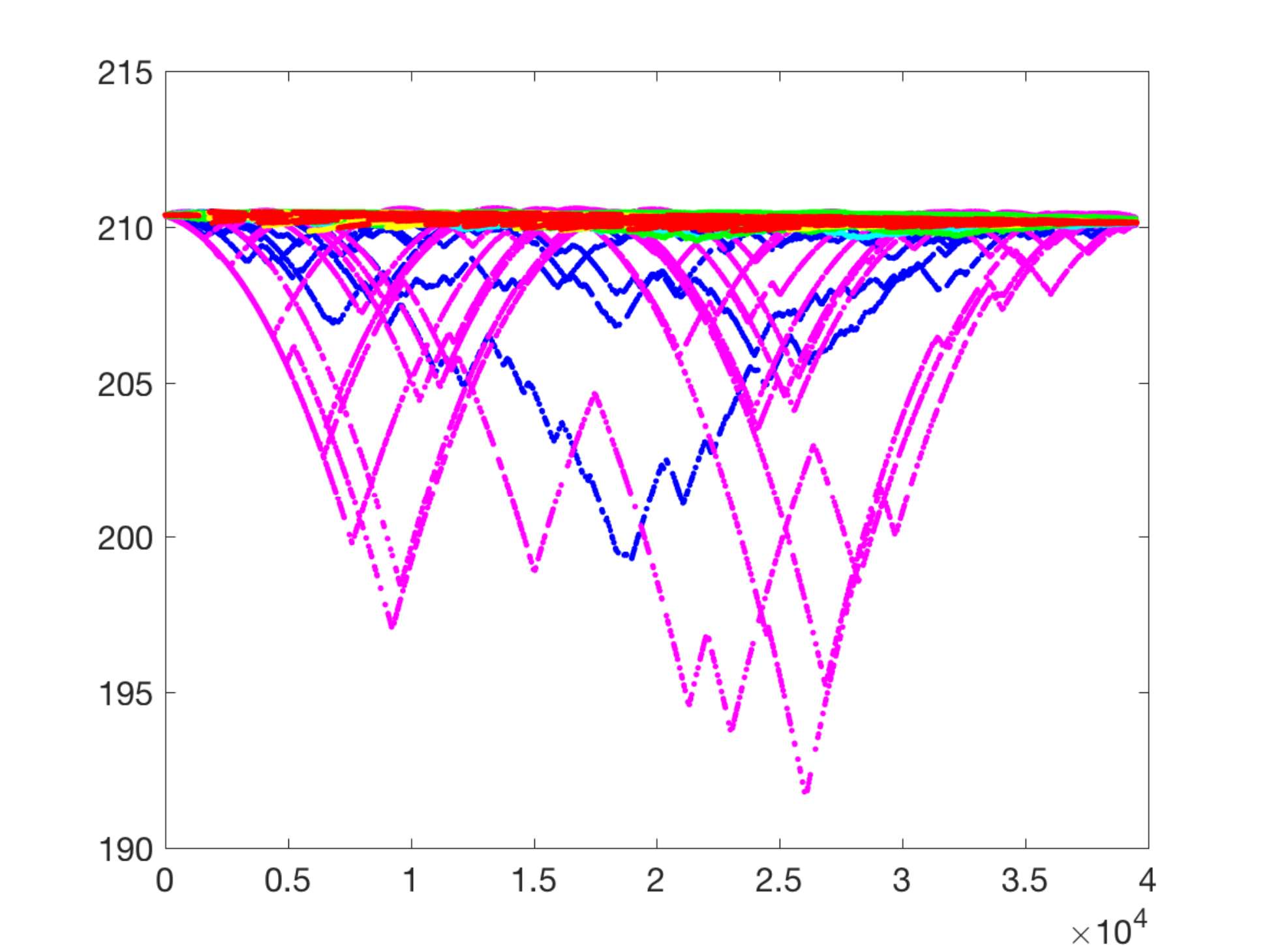} \hspace{-0.6cm}
        }

\caption{Features during transition for images for Asymmmetric Mutation  (\protect\plotblack), Uniform Random Walk  (\protect\plotnavy), Biased Random Walk (\protect\plotmagenta), EA-UniformWalk  (\protect\plotcyan), EA-BiasedWalk  (\protect\plotlime), EA-AsymUniformWalk  (\protect\plotyellow) and EA-AsymBiasedWalk  (\protect\plotrot) for images from Figure 1 (left), Black-White (middle), Figure~\ref{fig:color} (right). Generation number is shown on the $x$-axis and feature values on the $y$-axis.}

\label{fig:features}

\end{figure*}

Figure~\ref{fig:features} shows how the features evolve over time during the image transition process.
The first column refers to the transition process of the starting and target image given in Figure~\ref{fig:2}. The second column shows the transition of a complete black image starting image to a complete white target image, and the third column shows the transition of the color starting image to the color target image of Figure~\ref{fig:color}. Each figure shows the results of $10$ runs for each algorithm that we have considered for evolutionary image transition.

Considering the results for the images of Figure~\ref{fig:2} (left column), it can be observed that the feature values for Benford's Law reduce at the first half of the transition process and increase afterwards. Furthermore, the value for the target image is quite low, but the evolutionary image transition process produces images where the value for Benford's Law is significantly higher than the one for the starting and the target image in the last third of the image transition process. In respect to global contrast, it can also be observed that the transition process creates images of higher feature value than the ones of the starting and target image. All considered algorithms follow the same pattern for these two features, but it can be observed that the pure random walk algorithms of Section~\ref{sec3} overall achieve higher values for Benford's Law and the combined approaches are able to obtain a trajectory of higher values for Global Contrast Factor.

Considering the features Hue and Colorfulness, the feature values are following a more direct trajectory from the value of the starting image to the one of the target. For Hue, this trajectory is also very concentrated around the linear function connecting these two values whereas for Colorfulness a strong deviation, especially for the pure random walk algorithms of Section~\ref{sec3}, can be observed.

The transition process for the images of Figure~\ref{fig:color} carries out a process where the feature values of the starting and target image are of the same value. Again, it can be observed that the algorithms obtain higher values for Benford's Law and Global Contrast Factor during the transition for most of the runs. An exception is the biased random walk algorithm of Section~\ref{sec3} that sometimes produces lower values for these two figures during the transition. Hue and Colorfulness again exhibit a more direct trajectory between the starting and target feature value with the random walk algorithms showing a stronger fluctuation and in particular lower values with respect to Colorfulness.

Considering the transition for Black to White images, it can be observed that Benford's Law and Global Contrast Factor increase during the transition process. The concentrated behaviour for Benford's Law is due to the calculation of this feature as the feature value is fully determined by the number of black and white pixels. Furthermore, there are no changes during the transition process for Hue and Colorfulness.

The feature-base analysis allowed us to explore the different approaches for evolutionary image transition over time. Based on this, the user can obtain detailed insights into the behaviour of the seven approaches with respect to a given feature. 
As the different feature measurers are investigated on three different images, we can see the tendency of variations in the feature values for \emph{Benford's Law} and \emph{Global Contrast Factor} for all approaches throughout the transition process over time. In particular, the maximum value for \emph{Global Contrast Factor} obtained during the transition is significantly higher than the one for the starting and the target image. Interestingly, the feature values for \emph{Mean Hue} and \emph{Colorfulness}
exhibit a direct trajectory from the value of the starting image to the one of the target image.
In contrast, \emph{Colorfulness} shows a very strong fluctuation for the random walk algorithm during the whole transition process compared to the other approaches. In summary, the different aesthetic feature analysis gives the user valuable insights into the wide range of possibilities of using different evolutionary approaches for evolutionary image transition over time.

\section{Evolutionary Image Painting}
\label{sec7}

\begin{figure}[t] 
\centering

  \includegraphics[width=0.45\linewidth]{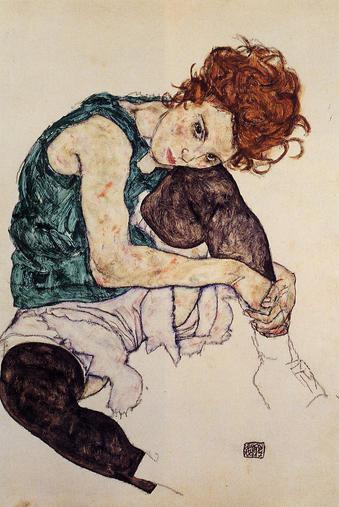}
  \includegraphics[width=0.45\linewidth]{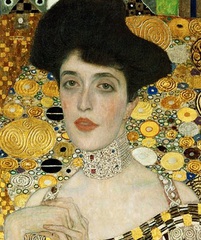}\\
\vspace{0.1cm}
\includegraphics[width=0.45\linewidth]{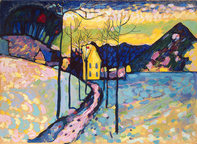}
\includegraphics[width=0.45\linewidth] {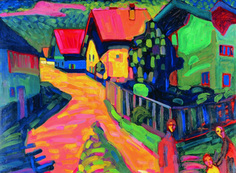}

\caption{Target images for evolutionary image painting.}
\label{fig:7X}
\end{figure}

\ignore{
\begin{figure*}[h]
\centering
     \includegraphics[width=0.2457\linewidth]{Images/painting_alpha_0_25k.jpg}
     \includegraphics[width=0.2457\linewidth]{Images/painting_alpha_0_5k.jpg} 
     \includegraphics[width=0.2457\linewidth]{Images/Painting_alpha_Nr1_0_75.jpg}
     \includegraphics[width=0.2457\linewidth]{Images/painting_alpha_1k.jpg} \\
        \vspace{0.3cm} 
         \includegraphics[width=0.2457\linewidth]{Images/Klimt_025k}
     \includegraphics[width=0.2457\linewidth]{Images/Klimt_05k} 
     \includegraphics[width=0.2457\linewidth]{Images/Klimt_075k}
     \includegraphics[width=0.2457\linewidth]{Images/Klimt_1k}

\caption{Evolutionary Image Painting with $\alpha=0.25$, $\alpha=0.5$, $\alpha=0.75$ and $\alpha=1$ with 12.5\%, 37.5\%, 62.5\% and 87.5\% of the target image (from left to right).}
\label{fig:7}

\end{figure*}  
}

We now consider how to use variations of the evolutionary image transition process for evolutionary image painting. The key idea is to make use of the biased random walk and use its behaviour of favouring similar colours. We use this property to change pixel values of a "blank" image such that it becomes similar to a given target image. A biased random walk mutation in the painting process uses for each step the same colour which is determined by the pixel of the target image where it started. Due to this, we call this process "painting" of the target image.

The evolutionary image painting algorithm is given in Algorithm~\ref{alg:5}. It is similar to the evolutionary transition algorithm and uses a starting image $S$ and the target image $T$ to be painted. Again, we assume $S_{ij} \not =T_{ij}$ for all pixels as pixels with $S_{ij} =T_{ij}$ can be viewed as already painted.
The algorithm minimizes the number of pixels that the current image $X$ agrees with $S$ and the pixels of $X$ with $X_{ij}=S_{ij}$ can be viewed as pixels that have not yet been visted.
 For our investigations, we use an all white starting image $S$ as we are mainly interested in the final painted image.

The mutation operator uses the biased random walk for a given starting pixel $X_{ij}$ (see Algorithm~\ref{alg:6}). As we are considering painting of an image, the starting pixel of the biased random walk determines the color in which the random walk paints the part of the image that it is visiting. The mutation operator in the evolutionary painting algorithm uses this biased painting random walk for each pixel that is still in the starting state $S_{ij}$ with probability $\min\left\{c_s/(2|X|_S),1 \right\}$ and therefore adapting the asymmetric mutation operator. It ensures that only biased random walks are started at pixels that have not changed their state yet. The idea is that pixels that have changed their state are considered as being painted and should therefore not change their color again.

If a biased random walk is started at a pixel $X_{ij}$, the color $C:=T_{ij}$ to be used during this random walk is given by the value of the pixel $T_{ij}$ in the target image $T$. During the biased random walk only pixels that have not changed their value yet are painted with the color $C$ which is again motivated by not painting pixels of the image that have already been painted. Each iteration of Algorithm~\ref{alg:5} does not increase the number of pixels where $X$ and $S$ agree. If at least one biased random walk happens, an offspring $Y$ with $f(Y,S) < f(X,S)$ (see definition of $f$ in Section~\ref{sec2}) is obtained. The implies that the algorithm minimizes the fitness function $f(X,S)$ and an image $X^*$ with $f(X^*,S)=0$ is considered to be fully painted. 

\begin{algorithm}[t]
\vspace{0.1cm}
\begin{itemize} 

\item Let $S$ be the starting image and $T$ be the target image.
\item Set $X:=S$.

\item while (not termination condition)
\begin{itemize}
\item $Y:=X$.
\item For each $Y_{ij} \in Y$ with $(Y_{ij}==S_{ij})$.
\begin{itemize}
\item Do Y:=PaintMutation($Y_{ij}, Y, S, T, \alpha, \tmax$) with probability $\min\left\{c_s/(2|X|_S),1 \right\}$. 
\end{itemize}
\item Set $X:=Y$.
\end{itemize}

\end{itemize}
\caption{Evolutionary image painting}
\label{alg:5}
\vspace{-0.12cm}
\end{algorithm}

\begin{algorithm}[t]
 \vspace{0.1cm}
\begin{itemize}
\item Set $C:= T_{ij}$.
\item Set $X_{ij} := C$.
\item $c:=0$.
\item while ($c \leq \tmax$ )
\begin{itemize}
\item $c:=c+1$
\item Choose $X_{kl} \in N(X_{ij})$ according to probabilities $p(X_{kl},\alpha)$.
\item Set $i:=k$, $j:=l$.
\item If $(X_{ij} == S_{ij})$ then  $X_{ij}:=C$. 
\end{itemize}
\item Return $X$.
\end{itemize}
\label{alg:6}
\vspace{-0.12cm}
\caption{PaintMutation($X_{ij}$, $X$, $S$, $T$, $\alpha$, $\tmax$)}
\label{alg:6}
\end{algorithm}

We now introduce a novel mutation operator called painting mutation operator for image transition developed to one specific purpose, in particular to produce interesting artistic images imitating modernist movement in the art called expressionism. The position of the first pixel $X_{ij}$ is chosen exactly the same how in our previous mutation operators, randomly. The painting mutation operator imposes a transition of the current starting image $X$ to an area of the target image $T$. Note that the paint mutation operator given in Algorithm~\ref{alg:6} only paints a pixel $X_{ij}$ with the chosen colour $C$ if the considered pixel is in the state of the starting pixel. A different option would be to always paint with the colour $C$ irrespective of whether the pixel $X_{ij}$ is in the starting or target state. This would allow that pixels already painted with a colour are able to change their colour during the process.

In our approach, we choose randomly one pixel from starting image $S$ and using the \brw described in detail in Section 3.2. The transition process occurs from starting image $S$, in our case we decided the image is white, to the target image, our artistic image. Through the transition from white image to another artistic image we can clearly represent processes the occur during \brw mutation. It is also possible to choose two artistic images when the colors are opposite of the color theory and achieve interesting effects in respect to stronger contrast. Our investigation has additionally the goal to give adequate insight in to the evolutionary processes during the transition processes.

\subsection{Impact of choice of the $\alpha$ in evolutionary image painting}

We have developed the special case of \brw mutation algorithm with constant $\alpha$. The classical \rw is running with $\alpha=0$ without deliberately currying about the $\alpha$. On the assumption \brw with $\alpha>0$ the algorithm will with high probability perform along the edges of the similar colors as opposed to going over the edges to the colors with greater RGB values differences.

\begin{figure*}[tpbH]
\centering   

    \includegraphics[width=0.497\linewidth]{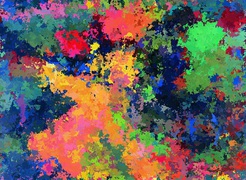} 
    \includegraphics[width=0.497\linewidth]{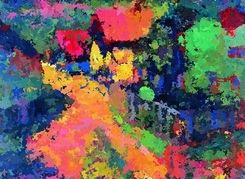}
    \hfil\\[0.3\tabcolsep] 
    \includegraphics[width=0.497\linewidth]{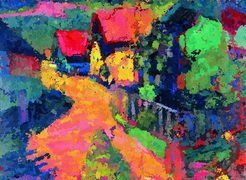} 
    \includegraphics[width=0.497\linewidth]{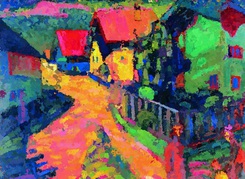}\\
    \vspace{0.3cm} 

    \includegraphics[width=0.497\linewidth]{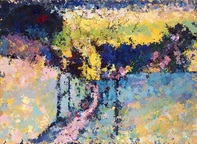} 
    \includegraphics[width=0.497\linewidth]{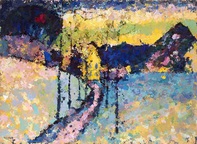}
    \hfil\\[0.2\tabcolsep] 
    \includegraphics[width=0.497\linewidth]{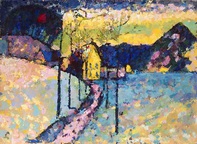} 
    \includegraphics[width=0.4965\linewidth]{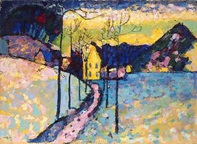}
 \vspace{0.1cm} 


\caption{Evolutionary Image Painting with
$t_{\max}=500$ and
$\alpha=0.25, 0.5, 0.75, 1.0$.
}
\label{fig:11}
\end{figure*}
\begin{sidewaysfigure}
   \centering
 

 \includegraphics[width=0.23\linewidth]{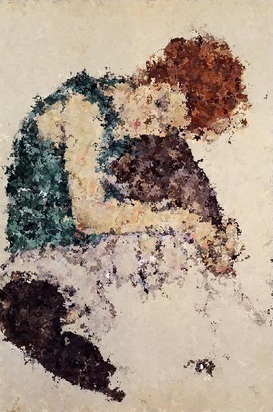} \quad
     \includegraphics[width=0.23\linewidth]{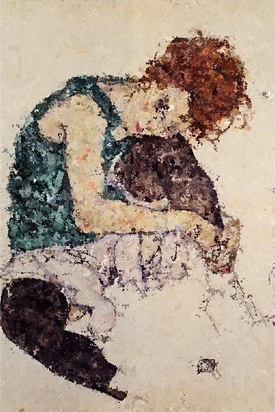} \quad
     \includegraphics[width=0.23\linewidth]{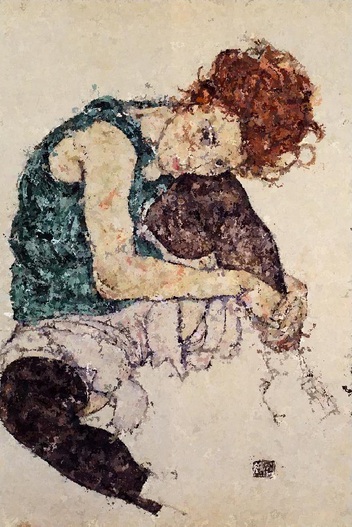} \quad
     \includegraphics[width=0.23\linewidth]{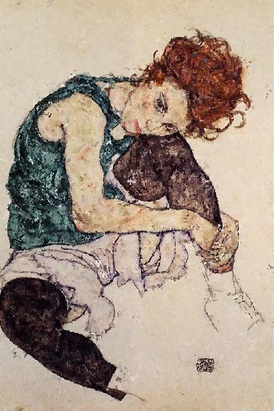} \quad
       \vspace{0.3cm} \\
         \includegraphics[width=0.23\linewidth]{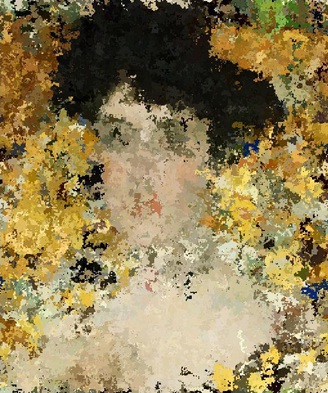} \quad
     \includegraphics[width=0.23\linewidth]{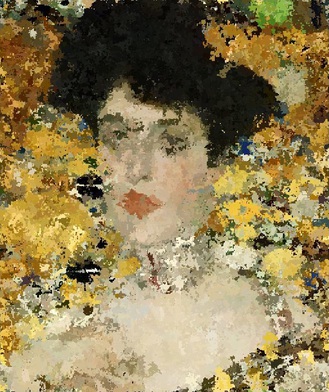} \quad
     \includegraphics[width=0.23\linewidth]{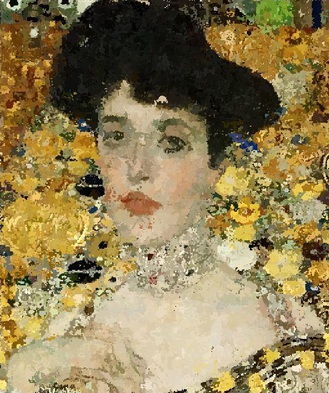} \quad
     \includegraphics[width=0.23\linewidth]{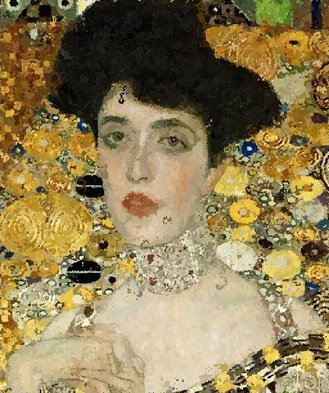} 
     
  
     \caption{Evolutionary Image Painting with $t_{\max}=500$ and $\alpha=0.25, 0.5, 0.75, 1$ (from left to right).}
\label{fig:12}

\end{sidewaysfigure}

For our experimental investigations, we set $c_s=200$ which implies that several random walks painting different parts of the image are started in each generation.
The goal of our experimental investigations is to study the effect of $\alpha$ in $p(X_{kl},\alpha)$ (introduced in Section~\ref{sec3}) for evolutionary image painting. This parameter allows to focus on similar color pixels during the biased random walk and one would expect a painted image close to the given target if $\alpha$ is large enough.

We study our evolutionary painting approaches on two landscape pictures and two portraits. The four target images used for evolutionary painting (see Figure~\ref{fig:7X}) are as follows: Seated woman with bent knee, 1917 by Egon Schiele; Adele Bloch-Bauer, 1907 by Gustav Klimt; Winter Landscape, 1909 by Wassily Kandinsky; Murnau street with women, 1908 by Wassily Kandinsky. These images have been selected  for our studies due to the wide recognition of these images as examples of fine art in Western culture~\citep{Iskin:2017}.
For our experimental investigations, we consider evolutionary painting with $\alpha =  0.25, 0.5, 0.75, 1.0$ and $\tmax = 500$. The results for evolutionary image painting with these different parameters of $\alpha$ are shown in Figure~\ref{fig:11} and Figure~\ref{fig:12}.

Our experiments show that the elements emerge over the image during the transition stage resulting in differently occurring images at the end of the generation processes. Figure~\ref{fig:11} and Figure~\ref{fig:12} show four various images with unique artistic value. Firstly, we see the target image $T$, following the four adjustment for $\alpha = 0.25, 0.5, 0.75, 1.0$. We have executed mostly varied appearance of the target image $T$. Each of these pictures represent different stages of the painting. Additionally, for comparison we have chosen $4$ different pictures as portray, landscape, abstract art and nature, in the stage of transition process when the painting process is completed. We can observe the impact on the evolutionary process cause throw the different settings.

As $\alpha$ controls the bias of the underlying random walks, it can be observed that small values of $\alpha$ obtain paintings that are much less precise than the target image. Thus, it's creating coarse grained painting effects. This effect decreases with increasing $\alpha$ and the painted image becomes much closer to the given target image. For all considered images, $\alpha=1$ already gives a painting being close to the target which is the reason why larger values of $\alpha$ are not investigated.
Comparing the different images, it can be observed that $\alpha=0.5, 0.75$ creates aesthetically pleasing paintings for the considered landscape pictures whereas a value of $\alpha=0.75$ produces novel results when applied to the considered portraits.

\subsection{Impact of Random Walk Length in Evolutionary Image Painting}

Now we investigate the impact of random walk length on our evolutionary painting approaches.
For our experimental investigations, we consider evolutionary painting algorithm with random walk length with $\tmax = 10, 100, 200, 1000, 4000, 10000$, and $\alpha = 0.25, 0.5, 0.75, 1.0$. We choose one of the landscape pictures presented in the Figure~\ref{fig:7X} and the abstract pictures presented in the Figure~\ref{fig:2}, what give us a clear comparison to the previous experiments described in Section \ref{sec7}. 

Also, we set $c_s = 200$ what again implies that several random walks painting different parts of the image are started in each generation. The goal of our experimental investigations is to study the effect of random walk length in $p(X_{kl},\alpha)$ introduced in Section \ref{sec3} for evolutionary image painting. This parameter allows to focus on similar color pixels during the biased random walk and one would expect a painted image close to the given target if α is large enough.

The results for evolutionary image painting with four different values of $\alpha$ and several lengths of random walk, which imply values of the range between large $\tmax = 10000$ and small $\tmax = 10$, are shown in Figure~\ref{fig:13} and Figure~\ref{fig:7_2alphaEIT_BRW}.

In the first row, we observe the image with $\tmax = 10$ and $\alpha = 0.25$, following the three adjustments for $\alpha = 0.5, 0.75, 1.0$. Each of these pictures represent different stages of the painting. Comparing the different images, it can be observed that $\alpha=0.25$, $\alpha=0.5 $ and $\tmax = 2000, 4000, 10000$ creates more patch-based apprentice of the image. We can observe the impact on the process cause throw the different settings of random walk length in evolutionary image paining.

\begin{figure*}[ht]
\centering

     \includegraphics[width=0.2419\linewidth]{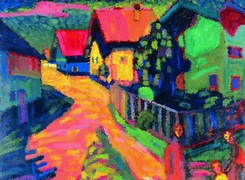}
     \includegraphics[width=0.2419\linewidth]{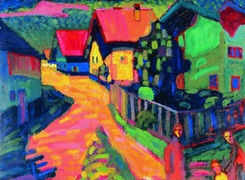}
     \includegraphics[width=0.2419\linewidth]{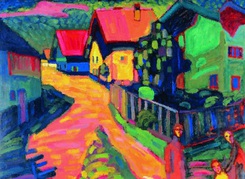} 
     \includegraphics[width=0.2419\linewidth]{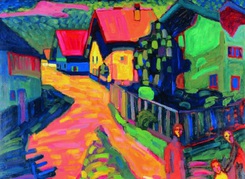}\\
      \vspace{0.3cm}
 \includegraphics[width=0.2419\linewidth]{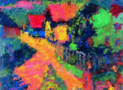} 
     \includegraphics[width=0.2419\linewidth]{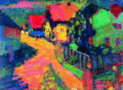}
     \includegraphics[width=0.2419\linewidth]{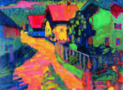} 
     \includegraphics[width=0.2419\linewidth]{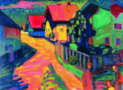}\\
      \vspace{0.3cm}
 \includegraphics[width=0.2419\linewidth]{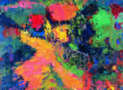}
     \includegraphics[width=0.2419\linewidth]{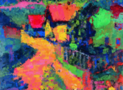}  
\includegraphics[width=0.2419\linewidth]{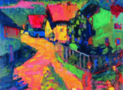}  
\includegraphics[width=0.2419\linewidth]{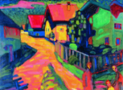}\\
       \vspace{0.3cm}
\includegraphics[width=0.2419\linewidth]{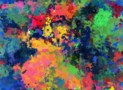}
     \includegraphics[width=0.2419\linewidth]{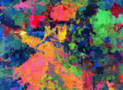}
     \includegraphics[width=0.2419\linewidth]{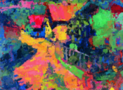} 
     \includegraphics[width=0.2419\linewidth]{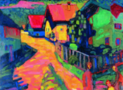}\\
      \vspace{0.3cm}
\includegraphics[width=0.2419\linewidth]{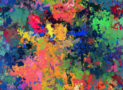}
     \includegraphics[width=0.2419\linewidth]{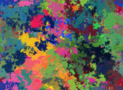}
     \includegraphics[width=0.2419\linewidth]{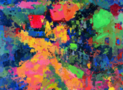} 
     \includegraphics[width=0.2419\linewidth]{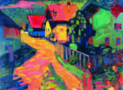}\\
      \vspace{0.3cm}
  \includegraphics[width=0.2419\linewidth]{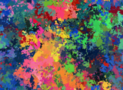}
     \includegraphics[width=0.2419\linewidth]{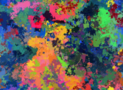}
     \includegraphics[width=0.2419\linewidth]{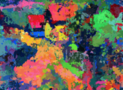} 
     \includegraphics[width=0.2419\linewidth]{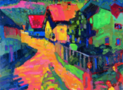}\\
      \vspace{0.3cm}
  \includegraphics[width=0.2419\linewidth]{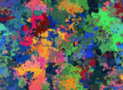}
     \includegraphics[width=0.2419\linewidth]{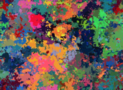}
     \includegraphics[width=0.2419\linewidth]{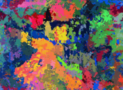} 
     \includegraphics[width=0.2419\linewidth]{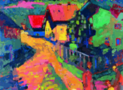}\\
      \vspace{0.3cm}
\caption{Evolutionary Image Painting with $\tmax=10, 100, 200, 1000, 2000, 4000, 10000$
(from top to bottom) and $\alpha=0.25, 0.5, 0.75, 1.5$ (from left to right).}
\label{fig:13}

\end{figure*}

In summary, the random walk length and the choice of $\alpha$ influences how "precisely" the target image is painted. The different parameter choices exhibit a wide range of possibilities for painting images with different effects.

\section{Conclusions and Future Work}

Evolutionary image transition uses the run of an evolutionary algorithm to transfer a starting image into a target image. In this paper, we have investigated how random walk algorithms can be used in the evolutionary image transition process. We have shown that mutation operators using different ways of incorporating uniform and biased random walks lead to different effects during the transition process. Furthermore, we have studied the impact of the different approaches with respect to different artistic features and observed that the process creates images which significantly differ from the starting and target image with respect to these features. 
Using biased random walks for image painting, we have shown that this approach creates artistic paintings which can be varied by the parameter $\alpha$ controlling the bias of the underlying random walk.

All our investigations are based on a fitness function that is equivalent to the well-known \ONEMAX problem. For future research, it would be interesting to study more complex fitness functions and their impact on the artistic behaviour of evolutionary image transition.

\section*{Acknowledgement}

This work has been supported through Australian Research Council (ARC) grant DP140103400.

\begin{figure*}[!ht]
\centering
     \includegraphics[width=0.23\linewidth]{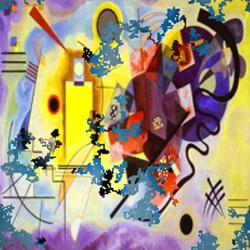}
     \includegraphics[width=0.23\linewidth]{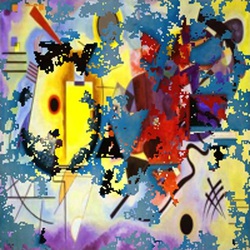}
     \includegraphics[width=0.23\linewidth]{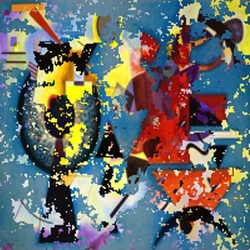} 
     \includegraphics[width=0.23\linewidth]{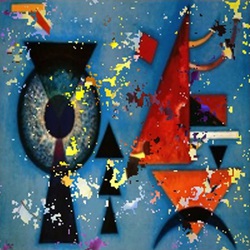}\\
         \includegraphics[width=0.23\linewidth]{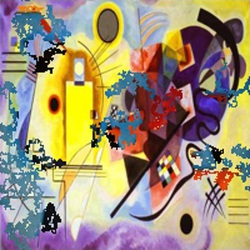} 
     \includegraphics[width=0.23\linewidth]{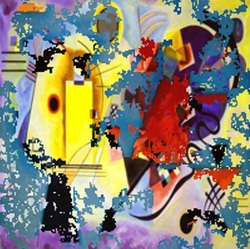}
     \includegraphics[width=0.23\linewidth]{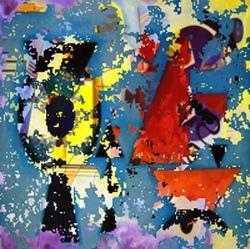} 
     \includegraphics[width=0.23\linewidth]{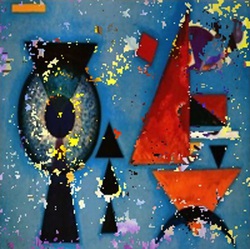}\\
     \includegraphics[width=0.23\linewidth]{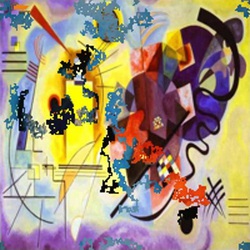}
     \includegraphics[width=0.23\linewidth]{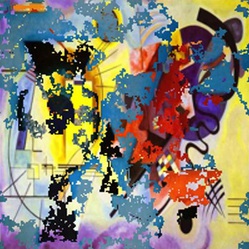}
     \includegraphics[width=0.23\linewidth]{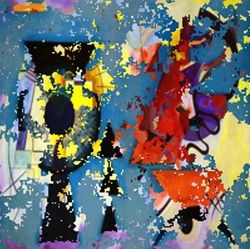}
     \includegraphics[width=0.23\linewidth]{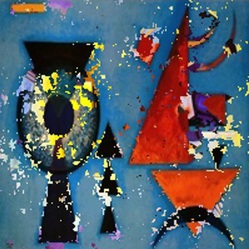}\\
     \includegraphics[width=0.23\linewidth]{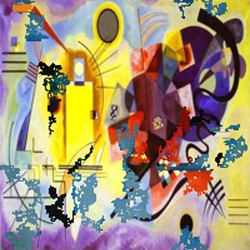}
     \includegraphics[width=0.23\linewidth]{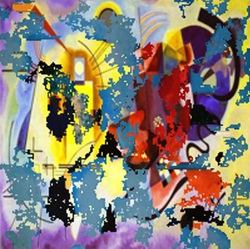}
     \includegraphics[width=0.23\linewidth]{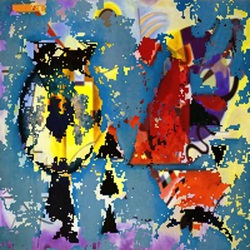} 
     \includegraphics[width=0.23\linewidth]{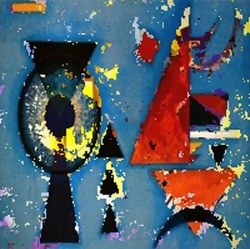}\\
           \includegraphics[width=0.23\linewidth]{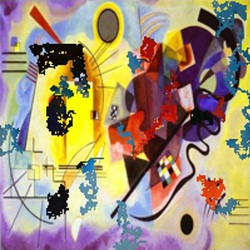}
     \includegraphics[width=0.23\linewidth]{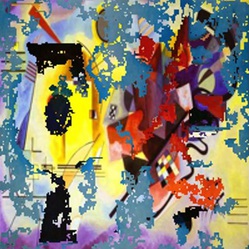}
     \includegraphics[width=0.23\linewidth]{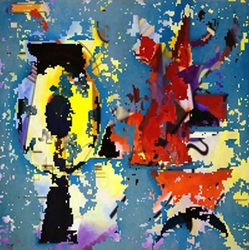} 
     \includegraphics[width=0.23\linewidth]{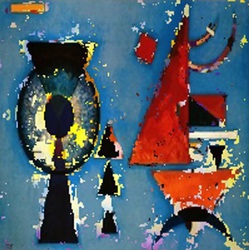}\\
           \includegraphics[width=0.23\linewidth]{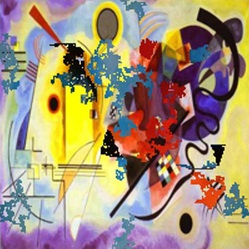}
     \includegraphics[width=0.23\linewidth]{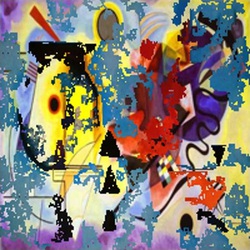}
     \includegraphics[width=0.23\linewidth]{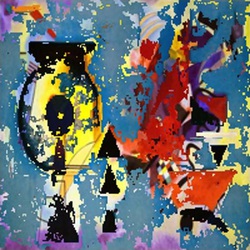} 
     \includegraphics[width=0.23\linewidth]{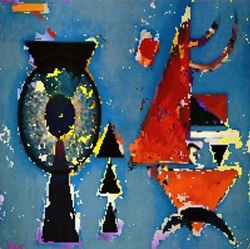}\\
           \includegraphics[width=0.23\linewidth]{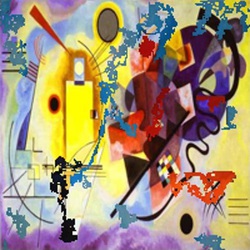}
     \includegraphics[width=0.23\linewidth]{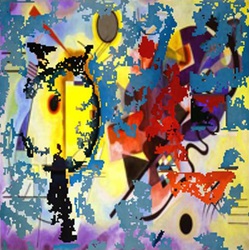}
     \includegraphics[width=0.23\linewidth]{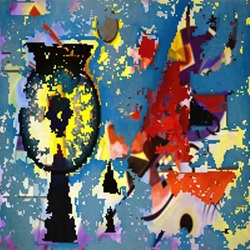} 
     \includegraphics[width=0.23\linewidth]{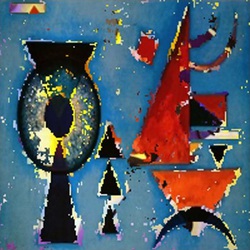}\\
      \vspace{0.3cm}
\caption{Evolutionary Image Transition with
$\alpha$ $  = 0.25, 0.5, 0.75, 1.0, 1.25, 1.5, 1.75$ 
with $12.5\%, 37.5\%, 62.5\%$ and $87.5\%$ of the target image (from top to bottom) and $\tmax=2000$.}   
\label{fig:7_2alphaEIT_BRW}
\end{figure*}

\bibliographystyle{apalike}

\bibliography{references}

\end{document}